\documentclass{article}

    \usepackage[final]{neurips_2025}

\usepackage[utf8]{inputenc} 
\usepackage[T1]{fontenc}    
\usepackage{hyperref}       
\usepackage{url}            
\usepackage[tracking=true]{microtype} 
\usepackage{amsmath, amsfonts}
\usepackage{booktabs}       
\usepackage{subcaption}     
\usepackage{amsfonts}       
\usepackage{nicefrac}       
\usepackage{microtype}      
\usepackage{multirow}
\usepackage{xcolor}         
\usepackage{makecell}       
\usepackage{caption}
\usepackage{enumitem}       
\usepackage{wrapfig}
\usepackage{graphicx}
\setlist[enumerate]{leftmargin=2em, labelsep=0.5em, label=\arabic*.}

\title{Towards Better \& Faster Autoregressive Image Generation: From the Perspective of Entropy}
\author{%
Xiaoxiao Ma$^{1,2}$ \quad
Feng Zhao$^{1\dagger}$ \quad
Pengyang Ling$^{1}$ \quad
Haibo Qiu$^{2}$\thanks{Project lead; $^{\dagger}$ Corresponding author.} \quad
Zhixiang Wei$^{1}$ \\[2pt]
\textbf{Hu Yu}$^{1}$ \quad
\textbf{Jie Huang}$^{1}$ \quad
\textbf{Zhixiong Zeng}$^{2}$ \quad
\textbf{Lin Ma}$^{2}$ \\[2pt]
\normalsize
$^{1}$University of Science and Technology of China \quad
$^{2}$Meituan \\
\normalsize
}

\begin{document}

\maketitle

\vspace{-3mm}
\begin{figure}[h]
\setlength{\abovecaptionskip}{0.2cm}
\setlength{\belowcaptionskip}{0.2cm}
  \centering
   \includegraphics[width=1\linewidth]{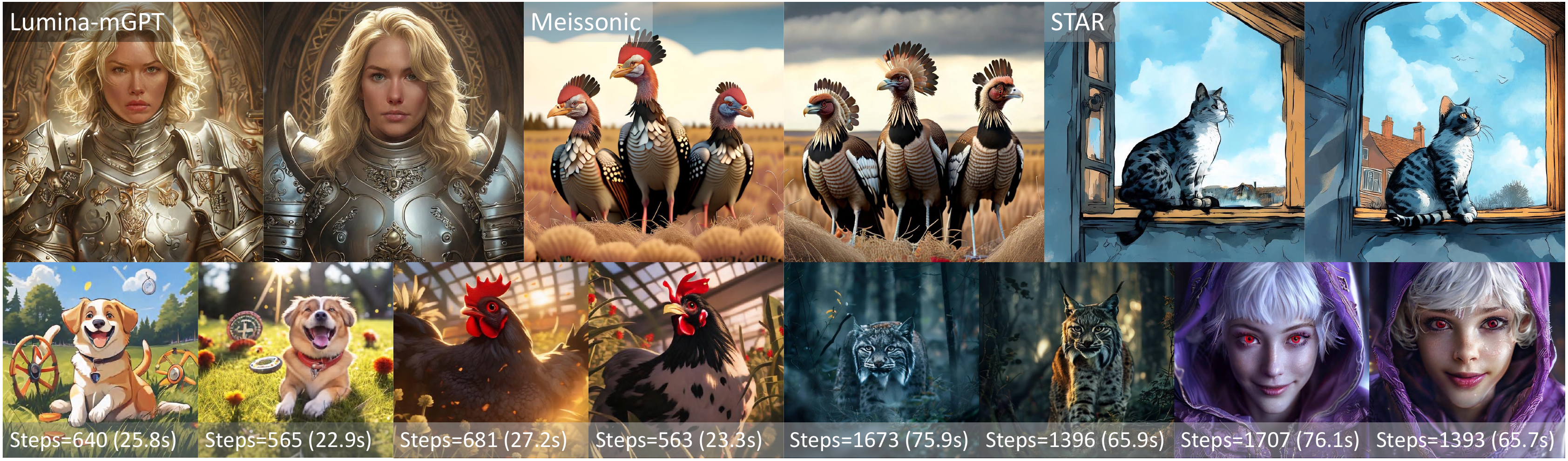}

   \caption{Top row: Our method generates images with finer details and better structure. Bottom row: Combined with existing acceleration methods, ours reduces inference cost by~15\%.
(Left two pairs are from LlamaGen~\cite{sun2024llamagen}; right from Lumina-mGPT~\cite{liu2024lumina_mgpt}. Inference steps and latency are reported.)}
   \label{fig:teaser}
\end{figure}
\begin{abstract}
In this work, we first revisit the sampling issues in current autoregressive (AR) image generation models and identify that image tokens, unlike text tokens, exhibit lower information density and non-uniform spatial distribution. Accordingly, we present an entropy-informed decoding strategy that facilitates higher autoregressive generation quality with faster synthesis speed.
Specifically, the proposed method introduces two main innovations: 1) dynamic temperature control guided by spatial entropy of token distributions, enhancing the balance between content diversity, alignment accuracy, and structural coherence in both mask-based and scale-wise models, without extra computational overhead, and 2) entropy-aware acceptance rules in speculative decoding, achieving near-lossless generation at about 85\% of the inference cost of conventional acceleration methods.
Extensive experiments across multiple benchmarks using diverse AR image generation models demonstrate the effectiveness and generalizability of our approach in enhancing both generation quality and sampling speed. Code is available at \textcolor{cyan}{\url{https://github.com/krennic999/ARsample}}.
\end{abstract}
\section{Introduction}
\label{sec_introduction}
Autoregressive (AR) modeling, as the mainstream in language generation~\cite{touvron2023llama,achiam2023gpt,touvron2023llama2,yang2024qwen2technicalreport}, has recently demonstrate strong potential in visual generation~\cite{liu2024lumina_mgpt,sun2024llamagen}, offering improved scalability~\cite{tian2024var} and potential for unified vision-language modeling~\cite{wu2024janus,chen2025janus_pro}.
In this paradigm, images are first quantized into discrete token sequences~\cite{esser2021vqgan,yu2021vit_vqgan}, which are then generated either token-by-token in a raster-scan order~\cite{wu2024janus,chameleonteam2025chameleon}, or in parallel through multi-token generation strategies~\cite{tian2024var,ma2024star,chang2022maskgit}.

Unlike diffusion or flow models that regress continuous tokens ~\cite{rombach2022stablediffusion,peebles2023dit,esser2024sdv3}, AR models learn the probabilistic distribution over discrete vocabulary, and sampling strategies (e.g., top-$K$, top-$p$~\cite{holtzman2019top_p}) are required to specify a token, which is essential and significantly impacts the quality and characteristics of generated content.
Within the area of language modeling, strategies have been proposed to augment reasoning capabilities and mitigate hallucinations, including logit shaping~\cite{tang2024top_nsigma,zhang2024edt}, contrastive decoding~\cite{su2022contrastive,chuang2023dola}, leveraging model-specific features ~\cite{huang2024opera}, and search-based methods~\cite{luo2024sed,guan2025rstar}, which emphasize answer accuracy over generation diversity.

However, a clear distinction exists between image and language: \textit{images exhibit lower information density and highly non-uniform spatial information distribution}, as shown in Fig.~\ref{fig_motivation} (a), making language-oriented methods suboptimal for image generation. This mismatch often leads to a trade-off between diversity in image contents and text consistency. As observed in~\cite{liu2024lumina_mgpt,teng2024sjd}, increasing randomness (\textit{e.g.,} high top-$K$) helps enrich visual content but compromises structural stability, leading to artifacts, distorted structures, or chaotic textures. Conversely, reducing randomness stabilizes structure and improves alignment, but often yields flat, oversmoothed details, or simplistic background.
How to balance randomness and determinism during sampling is thus critical for high-quality image generation. Unfortunately, existing methods typically rely on uniform sampling approaches like fixed top-$K$ or top-$p$, overlooking the inherent spatial information imbalance, which limits their ability to achieve high-quality images.

In this work, we aim to leverage the uneven distribution of information in images, and propose a \textit{sampling method specifically for autoregressive image generation}.
We observe that entropy of predicted logits effectively reflects information density in image during generation—low entropy corresponds to large homogeneous regions, while high entropy highlights content-rich areas such as foreground objects and complex backgrounds (see Fig.~\ref{fig_entropy_analysis}).
An intuitive idea is to encourage higher randomness in low-entropy regions, while applying stricter sampling in high-entropy areas. This helps balance image richness with structural stability, and also allocating fewer inference resources to low-entropy regions, which enables further acceleration with minimal impact on generation quality.
Unlike~\cite{wei2025_pure}, which applies entropy to control sampling randomness in super-resolution to modulate stochasticity, our work leverages entropy to guide autoregressive generation dynamics.

Building on this observation, we propose an entropy-aware sampling strategy that adjusts token distributions dynamically during inference. By computing the entropy of each predicted token distribution, we assign adaptive temperatures—injecting more randomness in low-entropy (simple) regions and applying stricter sampling in high-entropy (complex) areas. This improves the balance between image quality, structural stability, and text-image alignment without additional training or inference cost.
Moreover, our method generalizes well to a variety of autoregressive frameworks based on discrete token prediction, including mask-based and scale-wise generation. We also extend the entropy-aware idea to acceleration: by incorporating entropy-dependent acceptance in speculative decoding, we reduce inference cost to ~85\% of standard baselines with minimal quality loss.
We summarize our contributions as follows:

\begin{enumerate}
    \item Motivated by the observation that image information is sparse and unevenly distributed, which can be reflected by the entropy of tokens, we introduce an entropy-driven sampling strategy tailored for AR image generation that dynamically adapts sampling behavior based on entropy.
    \item In contrast to conventional sampling methods like top-$K$ or top-$p$, our approach enhances image quality and structural stability without modifying the model or increasing inference cost, and benefits multiple types of AR generation frameworks.
    \item We further extend the entropy-aware perspective to speculative decoding, achieving a 15\% reduction in inference time while maintaining visual fidelity across multiple benchmarks.
\end{enumerate}

\section{Related works}
\label{sec_related_works}
\subsection{Autoregressive image generation}
Early work~\cite{van2016pixelcnn} generates images directly at pixel level. Later approaches adopt a two-stage pipeline: images are first quantized into discrete tokens~\cite{esser2021vqgan,yu2021vit_vqgan}, then generated with Transformers in raster order~\cite{ding2021cogview,ge2023seed,ramesh2021dalle,yu2022parti,he2024mars,wang2024emu3}.
Recent efforts scale this paradign with larger models and stronger conditioning. LlamaGen~\cite{sun2024llamagen} provides class and text-conditioned baselines; Lumina-mGPT~\cite{liu2024lumina_mgpt} and Anole~\cite{chern2024anole} fine-tune Chameleon~\cite{chameleonteam2025chameleon} for improved text-conditioned generation. Unified frameworks further bridge understanding and generation~\cite{wu2024janus,chen2025janus_pro,jiao2025unitoken,unitok} in a single Transformer.
Meanwhile, image tokenizers have evolved for better reconstruction~\cite{yu2024titok,lee2022rqvae,yu2023lfq,zhao2024bsq} or multimodal integration~\cite{zhang2025v2flow,qu2024tokenflow}.

While proven effective, the vanilla autoregressive paradigm suffers from slow and rigid next-token prediction. To improve efficiency, recent studies explore more strategies, including multi-token prediction via random masking~\cite{chang2022maskgit,bai2024meissonic,xie2024show_o,yu2025videomar}, coarse-to-fine modeling~\cite{tian2024var,ma2024star,tang2024hart,han2024infinity,yu2025frequency}or hybrid approaches~\cite{he2025nar,yu2024rar}. Nonetheless, vector-quantized models still rely on sampling from token distributions, making generation quality sensitive to the sampling strategy.

\subsection{Sampling strategies in autoregressive models}
Transformers model the probability distribution over tokens, requiring specific sampling strategies to obtain concrete outputs. Common approaches in language modeling include top-\emph{k}~\cite{radford2019gpt2} and top-\emph{p}~\cite{holtzman2019top_p} sampling, which truncate the candidate space by rank or cumulative probability. EDT~\cite{zhang2024edt} dynamically adjusts temperature based on entropy to balance diversity and precision. Other approaches explore repetition penalties~\cite{keskar2019repetition_penalty}, contrastive decoding~\cite{chuang2023dola}, speculative decoding~\cite{leviathan2023speculative_decoding,chen2023accelerating}, and search-based techniques~\cite{meister2020beam_search,snell2024lookahead,guan2025rstar,lightman2023letsverifystepstep,snell2024scalingllmtts} to reduce hallucination or speed up inference.

In visual generation, a higher degree of randomness is often needed to produce more realistic and detailed content. LlamaGen~\cite{sun2024llamagen} and Lumina-mGPT~\cite{liu2024lumina_mgpt} demonstrate that much larger top-\emph{k} values than those used in language models help avoid over-smoothed and low-detail outputs. Recent methods~\cite{teng2024sjd,jang2025lantern} apply speculative~\cite{leviathan2023spec_decode} or parallel decoding~\cite{he2024zipar,wang2024par} to accelerate image synthesis. 
PURE~\cite{wei2025_pure} designs a top-\emph{k} strategy based on token entropy and detail levels to improve autoregressive super-resolution.
However, they overlook the highly uneven spatial information distribution in images during generation, and do not tailor decoding for autoregressive image generation.
\section{Methods}
\label{sec_methods}

In this section, we first introduce basic of autoregressive image generation in Sec.~\ref{sec_motivation}. Starting from the difference between image and text generation, we present our method from an entropy-based perspective by adjusting token-level randomness during generation (Sec.~\ref{sec_entropy_temperature}, Sec.~\ref{sec_adaption_more_ar}), and further extend this view to acceleration (Sec.~\ref{sec_ar_acceleration}).
\begin{figure*}[t]
\setlength{\abovecaptionskip}{0.1cm}
\setlength{\belowcaptionskip}{0.1cm}
\begin{center}
\includegraphics[width=1\textwidth]{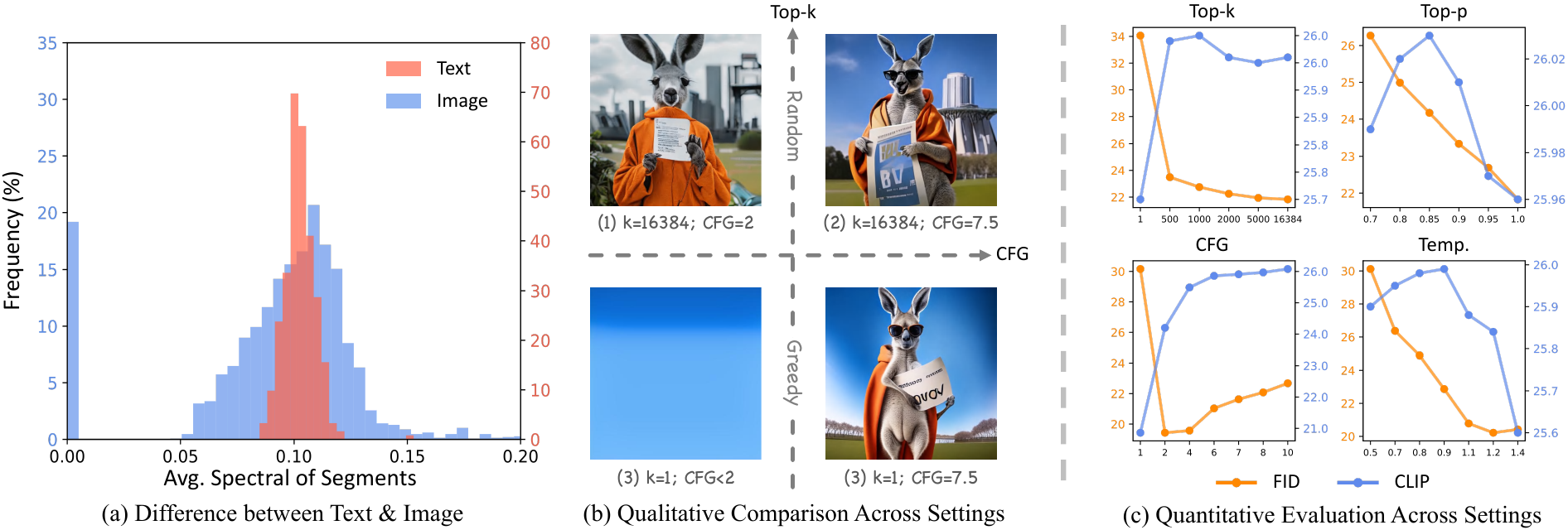}
\end{center}
\caption{
(a) Comparison of information density between image and text.
Histogram of average frequency-domain embeddings from LlamaGen~\cite{sun2024llamagen} (image) and Qwen2~\cite{yang2024qwen2technicalreport} (text) show the uneven spatial distribution in images with a large amount of low-frequency components.
(b) Qualitative results under various configurations. High CFG (Classfier-Free Guidance) or low top-$K$ often harms fidelity, while lower CFG with higher top-$K$ improves fidelity but may reduce text-image consistency.
(c) Quantitative evaluation of LlamaGen under different sampling settings. 
}
\vspace{-0.3cm}
\label{fig_motivation}
\end{figure*}

\subsection{Preliminaries \& motivation}
\label{sec_motivation}
\noindent\textbf{Autoregressive image generation.}
In a typical autoregressive generation process, an image $I\in \mathbb{R}^{H\times W\times 3}$ is quantized into a set of discrete tokens $(x_1, x_2, ..., x_{h\times w})$, where each token $x_i \in [V]$, $V$ denotes the size of the VQ-VAE codebook. The image tokens are generated sequentially by a transformer, with the $i+1$-th token $x_{i+1}$ conditioned on the previously generated tokens. This process is modeled as $\prod_{i=1}^{hw-1} p(x_{i+1} \mid x_{1:i})$, where $x_{1:i} = (x_1, x_2, ..., x_{i})$, and $p(x \mid x_{1:i})$ represents a categorical distribution over token at position $i+1$.
In text-conditioned generation tasks, the full image sequence is generated conditioned on a prefix of text tokens.
At each step, a sampling method such as top-$K$ or top-$p$ is applied to select a token from $p(x \mid x_{1:i})$. And choice of sampling strategy can significantly affect the quality of generated image, as illustrated in Fig.~\ref{fig_motivation} (b).

\begin{figure*}[t]
\setlength{\abovecaptionskip}{0.1cm}
\setlength{\belowcaptionskip}{0.1cm}
\begin{center}
\includegraphics[width=1\textwidth]{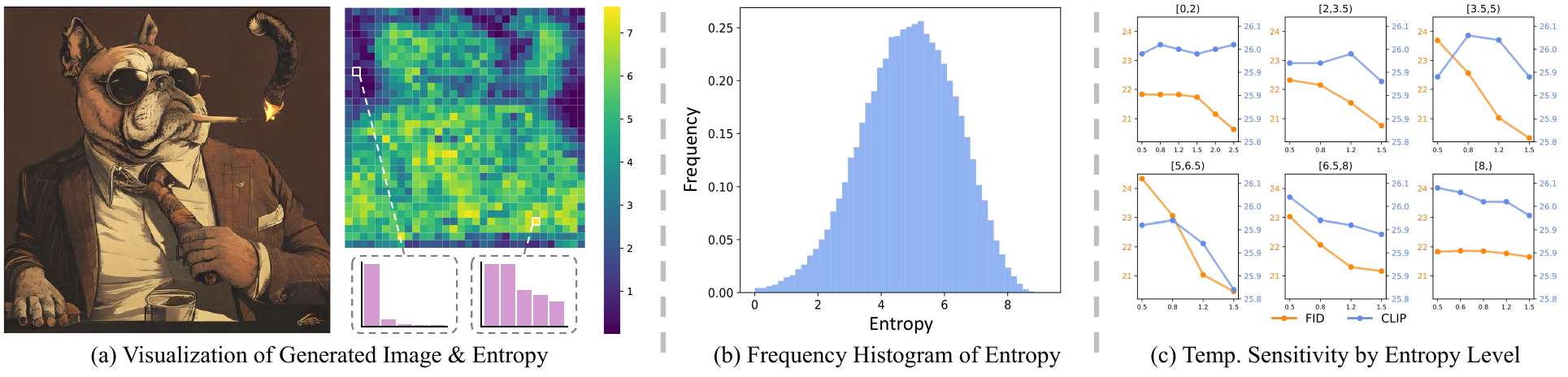}
\end{center}
\caption{
(a) Entropy map during generation: complex regions exhibit higher entropy (more dispersed probabilities), while simpler areas show lower entropy.
(b) Histogram of entropy distribution on COCO val2017 (from LlamaGen Stage II).
(c) Varying temperature by entropy range affects FID and CLIP score: lower-entropy tokens benefit from higher temperatures, and vice versa.
}
\vspace{-0.3cm}
\label{fig_entropy_analysis}
\end{figure*}

\noindent\textbf{Difference between image $\&$ text generation.}
Compared to text, \textit{the information density in images is lower and highly non-uniform}. For example, images often contain large regions of solid color or visually similar content, while nearby tokens in text are typically distinct. To illustrate the difference, we segment the embedding sequences of both images and texts into equal lengths, compute the average frequency spectrum per segment, and visualize the distributions in Fig.~\ref{fig_motivation} (a). The average frequency spectrum distribution of image segments is more dispersed, with a large amount of low-frequency areas; whereas textual segments demonstrate more compact and uniform distributions.

This discrepancy poses a challenge for sampling: fixed parameters like top-$K$ or top-$p$, though effective in language generation, perform suboptimally when simply applying them across all image tokens. They fail to account for spatial variability, resulting in regional artifacts: overly deterministic sampling may lose fine details, producing flat regions, excessive smoothness, or simplistic backgrounds; while excessively random sampling compromises semantic consistency and structural coherence, causing artifacts, distorted limbs, or chaotic textures.

\noindent\textbf{Relationship between entropy $\&$ image contents.}
We demonstrate that the \textit{entropy of predicted token distribution serves as an effective indicator of local information density} in an image.
Specifically, we compute the entropy $\epsilon$ of log-likelihood over all $V$ codebook entries at each generation step as:
\begin{equation}
\label{eq_entropy}
    \epsilon = - \sum_{k=1}^{V} p_k \log(p_k).
\end{equation}
As shown in Fig.~\ref{fig_entropy_analysis}, regions with simple content (\textit{e.g.,} solid colors) typically exhibit lower entropy, while more complex foreground (\textit{e.g.,} objects, structures, and textures) areas have higher entropy.
Low-entropy regions correspond to peaked distributions over a few tokens, indicating high model confidence. Conversely, high-entropy regions display more uniform distribution, reflecting greater uncertainty in token selection and higher information density. These observations validate entropy as a reliable proxy for measuring information density in images.

\subsection{Entropy-aware dynamic temperature}
\label{sec_entropy_temperature}
Building on the observation in Sec.~\ref{sec_motivation}, we further investigate how regions with different entropy levels affect image quality and the optimal sampling strategy.
Under a simple experimental setup, we adopt~\cite{sun2024llamagen} to analyze the entropy distribution of logits during generation, discretize it into intervals, and adjust token temperature within each interval to control sampling randomness. We then examine the relationship between image quality and text alignment via FID and CLIP-Score as indicators, results can be seen in Fig.~\ref{fig_entropy_analysis} (c). Our findings are as follows:
\begin{enumerate}
\item Across most entropy intervals, adjusting randomness leads to a trade-off between image quality and text alignment, especially for tokens in the entropy range of [2,8].
\item In high-entropy regions (\textgreater5), lower randomness helps improve text-image consistency.
\item In regions with extreme low entropy (\textless 2), increasing sampling randomness consistently improves visual quality, while having negligible impact on text alignment.
\end{enumerate}

These findings suggest that token-level sampling should be entropy-aware: during inference, tokens with \textit{lower entropy} should be assigned relatively \textit{higher randomness} to enhance the quality and visual richness of generated image, while \textit{high-entropy} tokens should be sampled \textit{more cautiously} to preserve clear structure, details and text alignment.

To better adapt to real-world inference, we introduce a dynamic temperature mechanism that adjusts sampling randomness on a per-token basis. Specifically, after computing the entropy of predicted distribution at each position, we determine a temperature value with predefined mapping function:
\begin{equation}
\label{eq_temperature}
    T=T_0 e^{-\frac{\epsilon}{\alpha}}+\theta,
\end{equation}
where $\epsilon$ denotes the entropy at current token position, $T_0$ represents the maximum temperature, $\theta$ sets the lower bound, and $\alpha$ controls the decay rate of temperature with increasing entropy. See Sec.~\ref{sec_ablation} for further analysis and discussion.
Subsequently, the resulting temperature $T$ is then applied to rescale the predicted logits as follows:
\begin{equation}
\label{eq_temperature2}
    \tilde{p}_i=\frac{p_i}{T}.
\end{equation}
Then, by applying $softmax(\tilde{p}_i)$, the differences between logits of different tokens are amplified (when $T$\textless 1) or reduced (when $T$\textgreater 1), which makes the probability distribution more concentrated or spread out, achieving the dynamic adjustment of sampling based on the region's content distribution.

\subsection{Adaptation to more AR models}
\label{sec_adaption_more_ar}
Additionally, many recent methods deviate from strict next-token prediction and instead adopt paradigms such as \textit{mask-prediction} or \textit{scale-wise} generation. We show that the proposed entropy-based strategy remains effective in these settings. After obtaining multiple token-level logit distributions from the transformer, we directly apply Eq.~(\ref{eq_temperature}) to them. As shown in Table~\ref{tab_results}, this approach consistently improves performance across standard evaluation metrics. Moreover, for different paradigms, specific designs can be incorporated to further enhance performance:

\begin{figure*}[t]
\setlength{\abovecaptionskip}{0.1cm}
\setlength{\belowcaptionskip}{0.1cm}
\begin{center}
\includegraphics[width=1\textwidth]{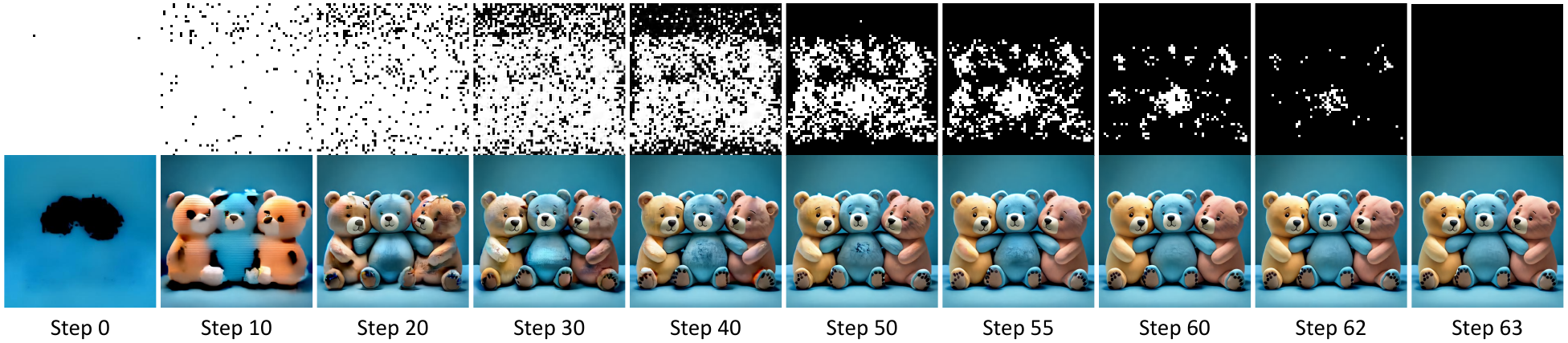}
\end{center}
\caption{
During generation of mask-based model~\cite{bai2024meissonic}, a large number of early steps (0$\sim$50) are allocated to compute tokens in simple regions, while only a few later steps (e.g., 50$\sim$63) for generating complex content. This often leads to degraded quality in the main visual subjects.
}
\vspace{-0.3cm}
\label{fig_visual_mask}
\end{figure*}

\noindent\textbf{Mask-prediction models.}
For mask-based models such as~\cite{bai2024meissonic,xie2024show_o}, a full probability distribution over all image tokens is obtained at each forward step. After sampling, a dynamic masking mechanism based on token confidence is applied. We observe that this masking strategy also has a significant impact on the quality of the generated results (see Fig.~\ref{fig_visual_mask}). Specifically, a soft categorical distribution is used to select $k$ tokens from all candidates to be accepted at the current timestep $t$:
\begin{equation}
\label{eq_mask_conf}
\text{conf} = \log p_t + T \cdot g,
\end{equation}
where $p$ is the predicted token probability, $T$ is the dynamic temperature defined in Eq. (\ref{eq_temperature}), $g \sim \text{Gumbel}(0, 1)$ is sampled from standard Gumbel distribution.
\begin{equation}
\label{eq_mask_conf2}
\mathcal{M}_t = \text{conf} < \text{TopK}(\text{conf}\odot \tilde{\mathcal{M}}_{t-1}, k).
\end{equation}
Here, $\text{TopK}(\text{conf}, k)$ returns the $k$-th highest confidence score, and tokens with lower confidence are masked out. $\mathcal{M}_t$ and $\mathcal{M}_{t-1}$ represent the accepted token masks at the current and previous timesteps, respectively, while ${\tilde{\mathcal{M}}_{t-1}}$ denotes the element-wise negation of $\mathcal{M}_{t-1}$. The number of accepted tokens $k$ at each timestep $t$ is determined by a predefined scheduler.
This design further encourages randomness in low-entropy regions while enhancing accuracy in high-entropy regions, leading to improved image quality.

\noindent\textbf{Scale-wise models}~\cite{tian2024var,ma2024star,tang2024hart,han2024infinity} generate tokens within each scale simultaneously. We find that assigning greater randomness to earlier scales while reducing randomness at later scales yields more accurate results without compromising image richness.
Specifically, we define a temperature term that decreases as the scale increases.
For the tokens at $s$-th scale, $T_s$ is calculated as:
\begin{subequations}
\begin{align}
T_s = T\cdot & [1-\beta \cdot (s-\left\lfloor S/2 \right\rfloor)];\\
& p_s = \frac{p_s}{T_s},
\end{align}
\end{subequations}
where $s$ denotes the scale index and $s \in \{1, 2, \dots, S\}$; $p_s$ is the logits of tokens at $s$-th scale, with shape of $h_s\times w_s \times V$. $T$ is dynamic temperature defined in Eq. (\ref{eq_temperature}). $\beta$ controls the decay rate of $T_s$ across scales and is set to 0.3 in experiments.

\begin{figure*}[t]
\setlength{\abovecaptionskip}{0.1cm}
\setlength{\belowcaptionskip}{0.1cm}
\begin{center}
\includegraphics[width=0.9\textwidth]{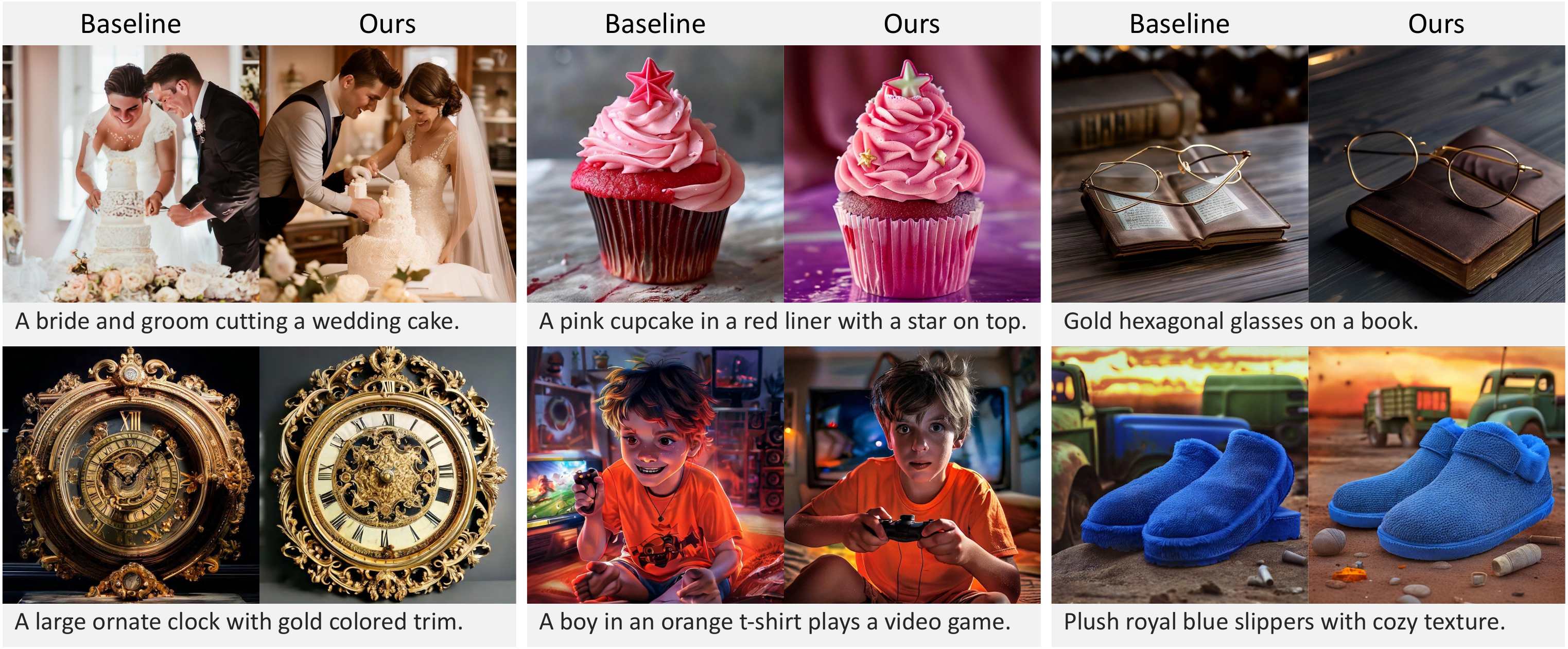}
\end{center}
\caption{
Visual comparison on next-token model.
Examples are from Lumina-mGPT, proposed method (``Ours'') maintains richer content while offering more accurate structure and finer details.
}
\vspace{-3mm}
\label{fig_results_mgpt}
\end{figure*}
\begin{table}[]
\centering
\setlength{\abovecaptionskip}{0.1cm}
\setlength{\belowcaptionskip}{0.1cm}
\caption{Performance of sampling strategies on various models.
``Baseline'' is original sampling; ``+Prob./Ours'' refers to Sec.~\ref{sec_entropy_temperature}; ``+Masking/+Scale-wise'' are paradigm-specific from Sec~\ref{sec_adaption_more_ar}.
}
\setlength{\tabcolsep}{4.5mm}{
\resizebox{0.9\columnwidth}{!}{%
\begin{tabular}{@{}ll|cccc@{}}
\toprule
Method                       & Config. & FID$\downarrow$   & CLIP-Score$\uparrow$ & DPG$\uparrow$ & HPSv2.1$\uparrow$ \\ \midrule
SDv2.1~\cite{rombach2022stablediffusion} & - & 22.87 & 26.31 & 68.09 & 26.38 \\
PixArt-$\alpha$~\cite{chen2023pixart_alpha} & - & 33.23 & 25.70 & 71.52 & 30.04 \\
SDXL~\cite{podell2023sdxl} & - & 23.20 & 26.46 & 74.21 & 28.54 \\
SDv3-medium~\cite{esser2024sdv3} & - & 29.82 & 26.24 & 85.85 & 30.22 \\ \midrule
\multirow{2}{*}{LlamaGen~\cite{sun2024llamagen}}    & Baseline    & 21.94 & 25.95 & 43.51 & 21.24  \\
                             & Ours        & \textbf{20.36} &  \textbf{25.96}  & \textbf{48.63} & \textbf{21.39}            \\ \midrule
\multirow{2}{*}{Lumina-mGPT~\cite{liu2024lumina_mgpt}} & Baseline    & 29.15 & 26.04     & 79.68 & \textbf{28.92}
\\
                             & Ours        & \textbf{27.44} & \textbf{26.25}      & \textbf{79.77} & 28.87  \\ \midrule
\multirow{3}{*}{Meissonic~\cite{bai2024meissonic}}   & Baseline    & 53.61 & 25.27      & 63.83 & 29.33   \\
                             & +Prob.      & \textbf{48.37} & 25.49      & 66.19 & 29.94   \\
                             & +Masking.   & 48.43 & \textbf{25.54}      & \textbf{67.08} & \textbf{30.04}   \\ \midrule
\multirow{3}{*}{STAR~\cite{ma2024star}}        & Baseline    & 35.05 & 25.43      & 70.25 & 28.79   \\
                             & +Prob.      & 32.75 & 25.56      & 70.83 & 28.93   \\
                             & +Scale-wise & \textbf{32.37} & \textbf{25.61}      & \textbf{70.86} & \textbf{29.06}   \\ \bottomrule
\end{tabular}
}}
\label{tab_results}
\vspace{-3mm}
\end{table}

\subsection{Autoregressive acceleration}
\label{sec_ar_acceleration}
We further explore the use of entropy to accelerate autoregressive generation.
Existing speculative decoding approaches~\cite{teng2024sjd,jang2025lantern} typically generate multiple candidate tokens via a draft model, followed by a verification step using a target model.
When the draft and target share the same model, the process reduces to comparing the confidence scores from two consecutive iterations. Specifically, the probability at the $(j{-}1)$-th step, $p(x \mid x_{1:i-1}^{(j-1)})$, and the $j$-th step, $p(x \mid x_{1:i-1}^{(j)})$, are compared. The acceptance probability of the token $x_i$ is then computed based on these two distributions:
\begin{equation}
p = \min \left( 1, \frac{p_{\theta}(x_{i}^{(j)} \mid x_{1:i-1}^{(j)})}{p_{\theta}(x_{i}^{(j-1)} \mid x_{1:i-1}^{(j-1)})} \right).
\end{equation}
In practice, a token is accepted if $p > r$, where $r$ is drawn from a uniform distribution $\mathcal{U}[0,1]$, which naturally balances randomness required for sampling diversity and accuracy during generation.

\begin{figure*}[t]
\setlength{\abovecaptionskip}{0.1cm}
\setlength{\belowcaptionskip}{0.1cm}
\begin{center}
\includegraphics[width=0.9\textwidth]{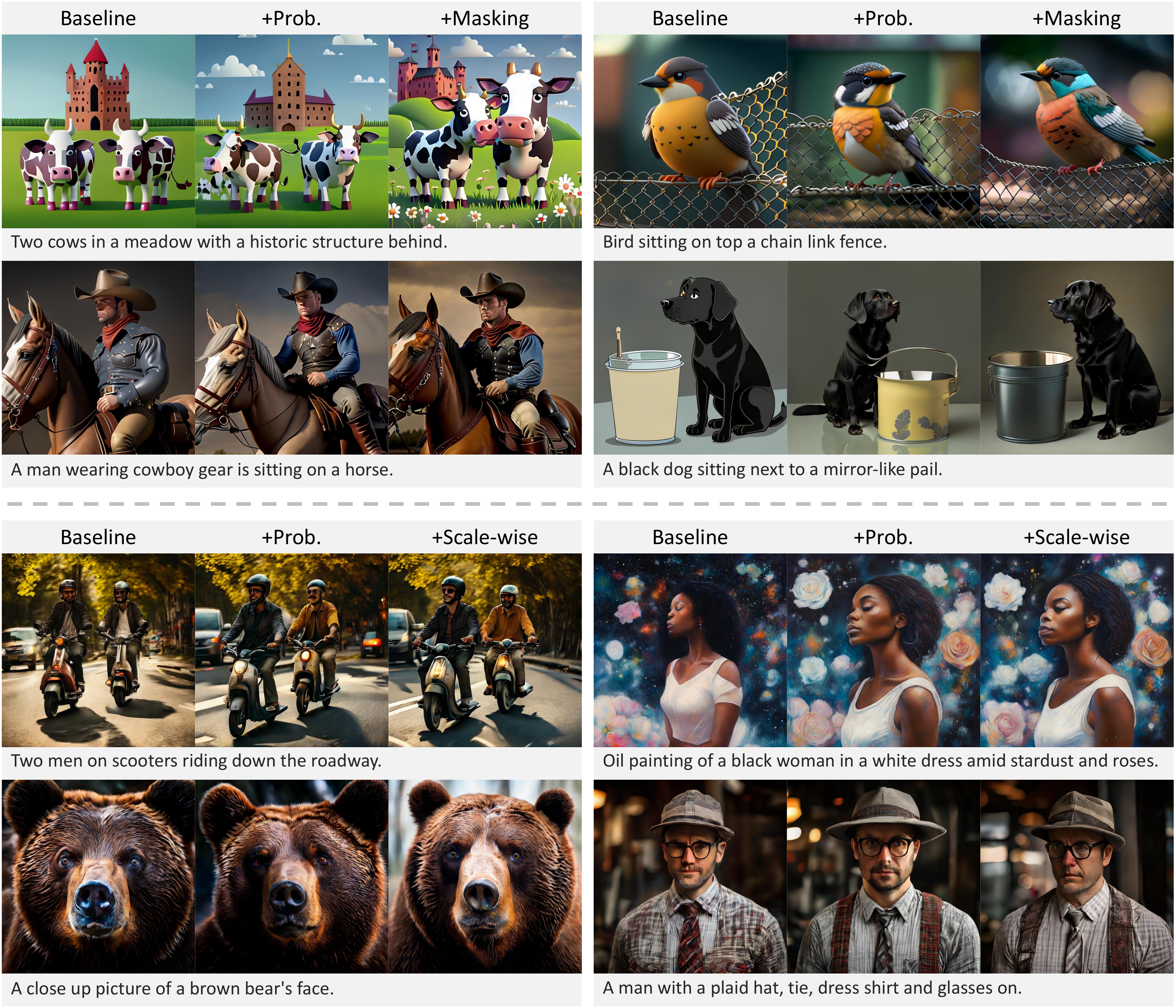}
\end{center}
\caption{
Visual comparison on mask-based (top) and scale-based model (bottom) from Meissonic and STAR. Proposed method provides better visual quality in structure and detail.
}
\vspace{-0.3cm}
\label{fig_results_star}
\end{figure*}

To make the process entropy-aware, we propose a simple modification to the acceptance rule. Since low-entropy regions are more predictable and allow higher randomness, while high-entropy regions require stricter verification, we scale the threshold $r$ by an entropy-based factor $\epsilon / e$, where $e$ is constant. This dynamic adjustment enables more efficient generation based on local uncertainty.

To improve stability, we rewrite $r$ as $0.5 + (r - 0.5)$, treating it as $0.5 + \mathcal{U}[-0.5, 0.5]$, where the noise term controls acceptance randomness. We scale this term with a decaying factor $(1 - \lambda \cdot \epsilon)$, where $\epsilon$ is the entropy, making high-entropy tokens more deterministically verified, while low-entropy areas retain near-uniform.
Combining both strategies above, the final acceptance rule can be formulated as:
\begin{equation}
p > \frac{\epsilon}{e} \left[ 0.5 + (r - 0.5)(1 - \lambda \cdot \epsilon) \right],
\end{equation}
where $\epsilon$ is entropy at the current token, $e$ and $\lambda$ are constants set to 8 and 16, respectively.

This dynamic acceptance criterion allocates inference budget more efficiently—being more permissive in confident regions and stricter in ambiguous ones, thereby reducing inference time with minimal performance loss. For detailed metrics and comparisons, please refer to Sec.~\ref{sec_exp_acc} and Sec.~\ref{sec_ablation}.
\section{Experiments}
\label{sec_experiments}
\subsection{Implementation details}
Four representative models are selected for comparison: vanilla AR model LlamaGen~\cite{sun2024llamagen} and Lumina-mGPT~\cite{liu2024lumina_mgpt} based on next-token prediction, mask-based model Meissonic~\cite{bai2024meissonic}, and scale-wise model STAR~\cite{ma2024star}.
All models are evaluated under their original inference settings (e.g., CFG=4 and top-$K$=2000 for Lumina-mGPT, CFG=7.5 for LlamaGen). We use LlamaGen’s official Stage-1 model to evaluate the sampling strategy, while Stage-2 is used only for acceleration analysis due to its poor performance (FID 53.42, CLIP-Score 21.47), which makes quality differences hard to observe.

FID and CLIP-Score are tested on the MS-COCO 2017~\cite{lin2014coco} validation set to evaluate the image quality and prompt-following capability. Moreover,  DPG-bench~\cite{hu2024ella} and HPS~\cite{wu2023hps} are adopted to assess the semantic fidelity and perceptual quality of the generated images. All experiments are conducted on A100 GPUs.

\subsection{Sampling quality}
\label{sec_exp_sample}
As shown in Table~\ref{tab_results} and Fig.~\ref{fig_results_mgpt}–\ref{fig_results_star}, our dynamic temperature sampling strategy effectively adapts to regions with varying information density in the image, leading to more stable structures and clearer details in the generated outputs. Depending on the inherent sampling mechanism of each model, our method yields varying degrees of improvement across different approaches. In particular, it achieves an approximate 4-point gain on DPG for both Meissonic and LlamaGen, along with a notable enhancement in visual quality.
In addition, integrating our approach with the masking- and scale-wise strategies described in Sec.~\ref{sec_adaption_more_ar} can further enhance generation performance. See Table~\ref{tab_results} for results with ``+Prob'' (applying dynamic temperature to logits only), and ``+Masking / +Scale-wise'' (applying temperature based on mask or scale).

\begin{figure*}[t]
\setlength{\abovecaptionskip}{0.1cm}
\setlength{\belowcaptionskip}{0.1cm}
\begin{center}
\includegraphics[width=0.95\textwidth]{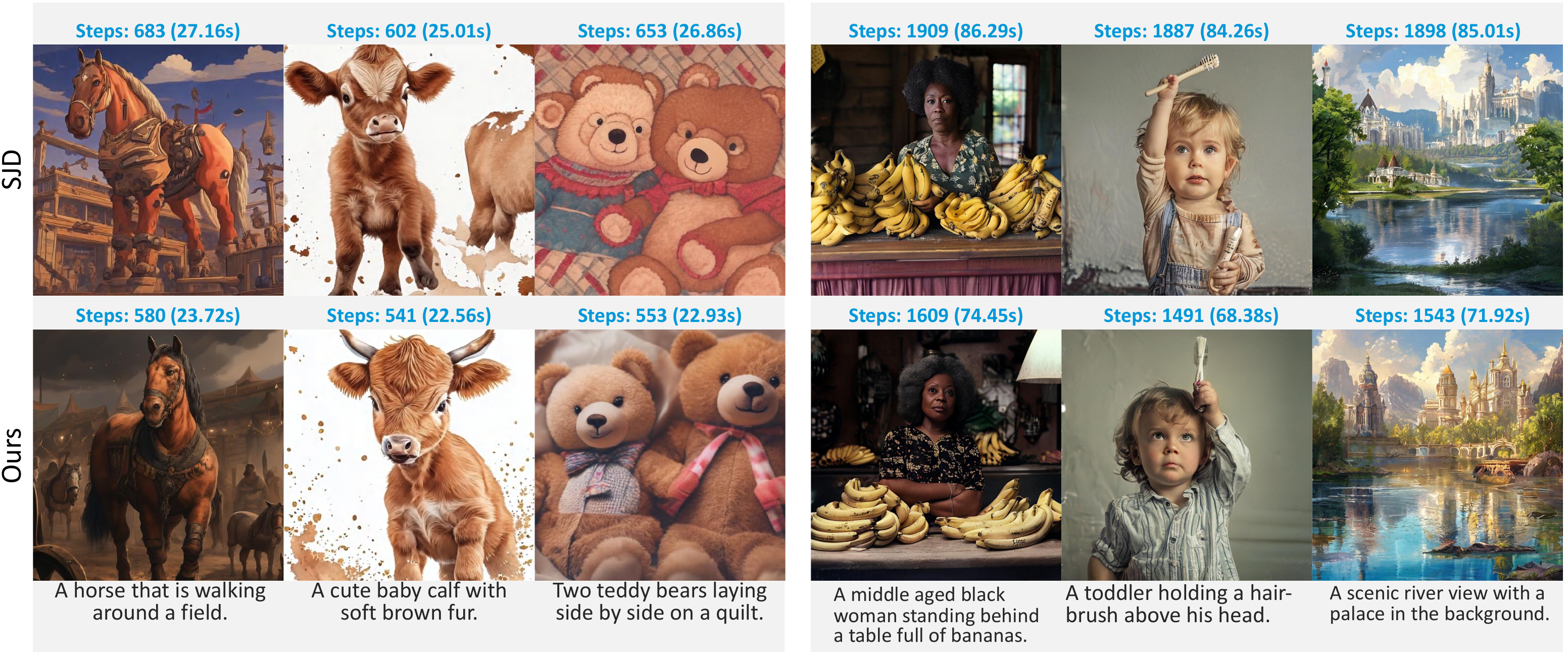}
\end{center}
\vspace{-1mm}
\caption{
Results on acceleration. We report the inference steps (``Steps'') and latency. Our method achieves similar image quality while using only 85\% of the baseline's (``SJD'') inference cost.
}
\vspace{-3mm}
\label{fig_results_acc}
\end{figure*}
\begin{table}[]
\centering
\setlength{\abovecaptionskip}{0.1cm}
\setlength{\belowcaptionskip}{0.1cm}
\caption{Quantitative comparison of acceleration on COCO17-val. ``Vanilla'' refers no acceleration.}
\setlength{\tabcolsep}{2.5mm}{
\resizebox{0.9\columnwidth}{!}{%
\begin{tabular}{@{}ll|cc|ccc@{}}
\toprule
Model & Config. & Avg. Latency [s]$\downarrow$ & Avg. Steps$\downarrow$ & FID$\downarrow$ & CLIP-Score$\uparrow$ \\ \midrule
\multirow{3}{*}{LlamaGen~\cite{sun2024llamagen}} &  Vanilla  & 44.52 & 1024 & 53.42 & 21.47         \\
       &  SJD~\cite{teng2024sjd} & 26.52  & 626.9 & 54.49 & 21.45         \\
       & Ours & \textbf{22.04}  & \textbf{535.5} & 54.60 & 21.51         \\ \midrule
\multirow{3}{*}{Lumina-mGPT~\cite{liu2024lumina_mgpt}} & Vanilla & 169.72 & 4165 & 29.15 & 26.04         \\
       & SJD~\cite{teng2024sjd}  & 84.97 & 1854.5 & 30.76 & 26.09         \\ 
       & Ours & \textbf{72.35} & \textbf{1594.5} & 30.89 & 26.10         \\ \bottomrule
\end{tabular}
}}
\label{tab_results_acc}
\vspace{-3mm}
\end{table}

\subsection{Inference acceleration}
\label{sec_exp_acc}
By integrating with existing vision-based speculative decoding schemes and leveraging entropy to automatically control the acceptance condition, our method saves about 15\% inference cost with almost no loss in image generation quality compared to the approach in~\cite{teng2024sjd}, as shown in Table~\ref{tab_results_acc} and Fig.~\ref{fig_results_acc}. The entropy-based approach significantly reduces the number of inference steps and latency, while still maintaining comparable image quality to the original speculative decoding method.

\subsection{Ablation \& discussion}
\label{sec_ablation}
\begin{figure*}[t]
\setlength{\abovecaptionskip}{0.1cm}
\setlength{\belowcaptionskip}{0.1cm}
\begin{center}
\includegraphics[width=1\textwidth]{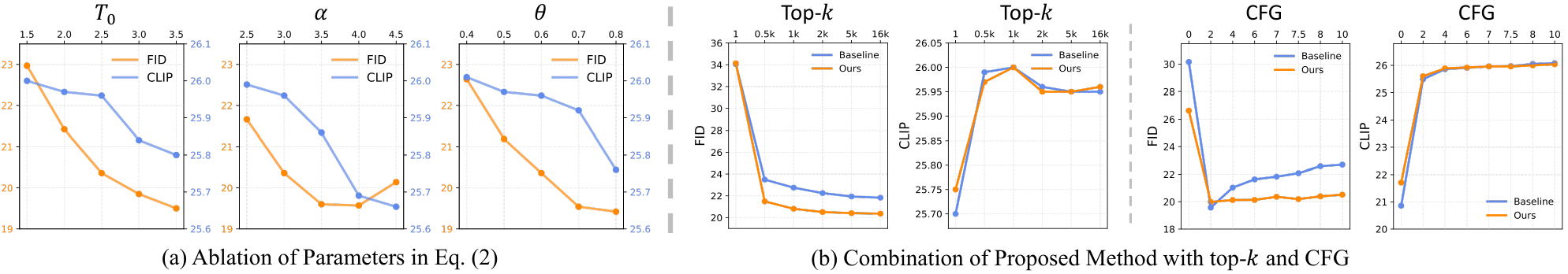}
\end{center}
\caption{
(a) Impact of parameters in Sec.~\ref{sec_entropy_temperature}; (b) Combination of our sampling strategy with existing methods (Top-$K$, CFG). Our method reduces the sensitivity of FID and CLIP-Score to these parameters, enhancing image quality and text alignment. Experiments are conducted on LlamaGen.
}
\vspace{-3mm}
\label{fig_ab_param}
\end{figure*}

\noindent\textbf{Parameters in Eq.~\ref{eq_temperature}.}
The impact of parameters in entropy-aware dynamic temperature is provided in Fig.~\ref{fig_ab_param}(a). It is observed that smaller $\epsilon$ or $\theta$ values lead to higher FID and lower CLIP-Score, primarily due to decreased randomness and overly deterministic sampling. Meanwhile, FID shows a trend of initially decreasing and then increasing as $\alpha$ increases. This is because $\alpha$ governs the proportion of different temperatures. When $\alpha$ is too small, most tokens are assigned very low temperatures, causing the FID to increase. Conversely, when $\alpha$ is too large, the image content becomes overly chaotic, resulting in increased FID. CLIP-Score consistently decreases as $\alpha$ increases.

\begin{figure*}[t]
\centering
\noindent
\begin{minipage}[t]{0.47\columnwidth}
\vspace{1mm}
\setlength{\abovecaptionskip}{0.1cm}
\setlength{\belowcaptionskip}{0.1cm}
\centering
  \centering
   \includegraphics[width=1\linewidth]{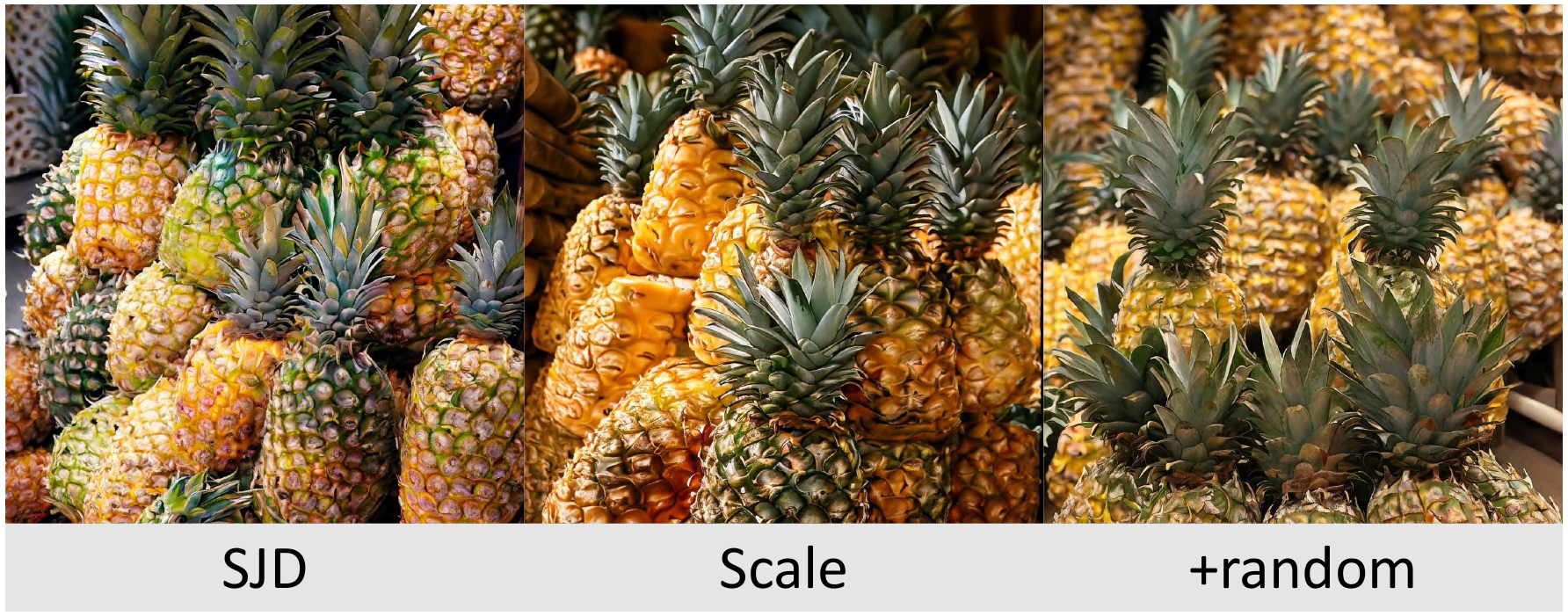}
   \captionof{figure}{Visual comparison for Table~\ref{tab_ab_acc}. ``Scale'' slightly degrades quality while accelerating, which can be mitigated by +random''.}
   \label{fig_ab_acc}
\end{minipage}%
\hfill
\begin{minipage}[t]{0.5\columnwidth}
\centering
\setlength{\abovecaptionskip}{0.1cm}
\setlength{\belowcaptionskip}{0.1cm}
\captionof{table}{Quantitative comparison on acceleration. ``Avg. Lat." is short for ``Avg. Latency.'' Experiments are conducted on Lumina-mGPT.}
\setlength{\tabcolsep}{3mm}{
\resizebox{1\columnwidth}{!}{%
\begin{tabular}{@{}l|llll@{}}
\toprule
Config. & \makecell{Avg.\\Lat.$\downarrow$} & \makecell{Avg.\\NFE$\downarrow$} & \makecell{FID$\downarrow$} & \makecell{CLIP\\Score$\uparrow$} \\ \midrule
SJD~\cite{teng2024sjd}           & 84.97        &  1854.5  & 30.76 & 26.09  \\
scale         & 69.38        &  1523.7  & 33.86 & 26.05  \\
+random   & 72.35       & 1594.5  & 30.89 & 26.12 \\
\bottomrule
\end{tabular}
}}
\label{tab_ab_acc}
\end{minipage}
\vspace{-3mm}
\end{figure*}

\noindent\textbf{Acceptance rate in Sec.~\ref{sec_ar_acceleration}.}
We propose to dynamically control the acceptance rate in existing speculative decoding methods based on the entropy of predicted distributions. By adjusting both the scale and randomness of the threshold $r$, we reduce latency while maintaining quality. As shown in Table~\ref{tab_ab_acc} and Fig.~\ref{fig_ab_acc}, controlling only scale of $r$ (``scale'') reduces inference cost but degrades performance, especially image quality. In contrast, jointly tuning both scale and randomness (``+random'') achieves a better trade-off, enabling high-quality generation with minimal inference overhead.

\noindent\textbf{Compatibility with different AR models.}
Our sampling method brings notable performance gains for some models—for instance, DPG in LlamaGen and Meissonic outperforms baseline by over 3 points. In contrast, well-trained models like Lumina-mGPT benefit only marginally. This discrepancy stems from factors such as generation paradigm (\textit{e.g.}, inherent sampling limitations of mask-based methods discussed in Fig.~\ref{fig_visual_mask}), training datasets and iterations (\textit{e.g.}, whether has been thoroughly trained on large-scale data). Nevertheless, these models can still exploit entropy for further acceleration.

\begin{figure*}[t]
\centering
\noindent
\begin{minipage}[t]{0.62\columnwidth}
\vspace{1mm}
\setlength{\abovecaptionskip}{0.1cm}
\setlength{\belowcaptionskip}{0.1cm}
\centering
  \centering
   \includegraphics[width=1.0\textwidth]{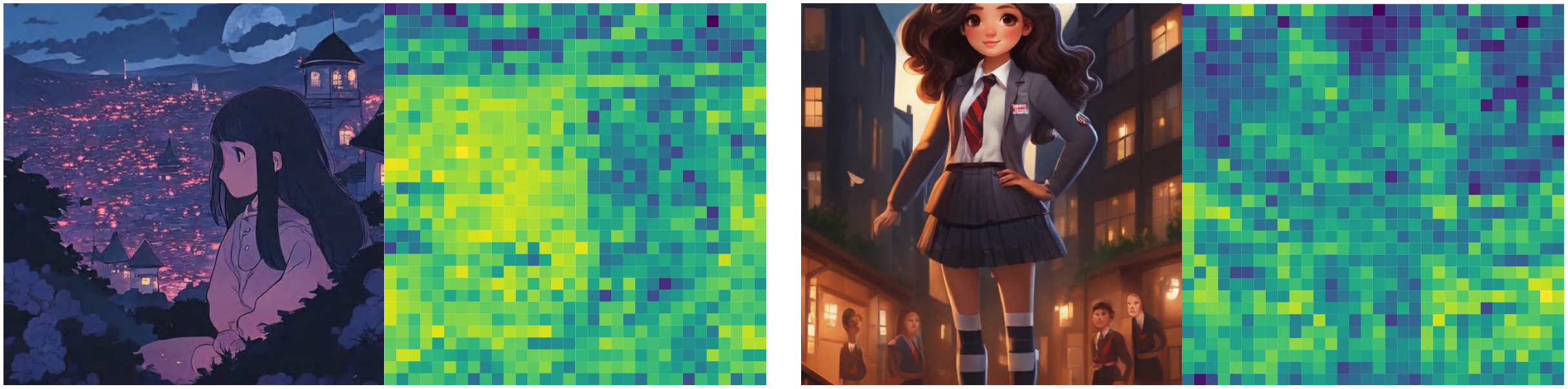}
   \captionof{figure}{For some cases, the semantic-corresponding foreground contents may have smaller entropy.}
   \label{fig_discuss_failurecase}
\end{minipage}%
\hfill
\begin{minipage}[t]{0.35\columnwidth}
\vspace{-2mm}
\centering
\setlength{\abovecaptionskip}{0.1cm}
\setlength{\belowcaptionskip}{0.1cm}
\captionof{table}{Mean and variance (``Var.") of entropy from~\cite{sun2024llamagen} on COCOval17.}
\setlength{\tabcolsep}{3.7mm}{
\resizebox{1.0\columnwidth}{!}{%
\begin{tabular}{@{}cc|cc@{}}
\toprule
Reso. & CFG & Mean & Var. \\ \midrule
256   & 7.5 & 4.76 & 2.75 \\
256   & 4.0 & 5.90 & 2.42 \\
256   & 2.0 & 6.87 & 1.97 \\
512   & 7.5 & 4.42 & 3.09 \\ \bottomrule
\end{tabular}
}}
\vspace{-2mm}
\label{tab_discuss_entropy}
\end{minipage}
\vspace{-3mm}
\end{figure*}

\noindent\textbf{Combination with top-$K$ and CFG.}
We further analyze the performance of our method when combined with top-$K$ sampling and CFG, as shown in Fig.~\ref{fig_ab_param}(b); results with top-$p$ and temperature are in the supplementary. By incorporating proposed method, FID metric becomes less sensitive to sampling parameters, enabling better fidelity while maintaining image-text alignment.

\noindent\textbf{Factors affecting entropy.} Unlike text generation with fixed tokenization rules, autoregressive image generation relies on pretrained tokenizers, and the underlying model differences—including parameter scale, data quality, and training corpus—lead to varying entropy distributions and optimal sampling parameters. Empirically, higher CFG and larger resolutions lead to lower average entropy (see Table~\ref{tab_discuss_entropy}). Further analyses are provided in supplementary material.

\noindent\textbf{Discussion of failure cases.}
Although our entropy-based dynamic sampling strategy brings notable performance improvements, we also observe several failure cases where the relationship between semantic information and the entropy map becomes less consistent (see Fig.~\ref{fig_discuss_failurecase}). In some cases, regions such as human faces exhibit unexpectedly high entropy, while complex backgrounds receive lower entropy values. Consequently, adjusting temperature based on such entropy patterns may lead to structural distortions and overly smooth details. This ambiguity may potentially limit further performance gains, especially for models that have been carefully optimized.
\section{Conclusion}

In this work, we first point out the need for different sampling strategies in autoregressive image and text generation, given their distinct information distributions. 
Starting from this perspective, we find entropy effectively represents image information density, offering new possibilities for improving and accelerating image generation.
As our method involve parameter adjustments without training, this approach could be further integrated into training or fine-tuning frameworks, potentially accelerating training, boosting inference speed, improving stability, and reducing hyperparameter dependence.

\clearpage
{
\small
\bibliographystyle{unsrt}
\bibliography{main}
}

\clearpage
\newpage
\section*{NeurIPS Paper Checklist}

\begin{enumerate}

\item {\bf Claims}
    \item[] Question: Do the main claims made in the abstract and introduction accurately reflect the paper's contributions and scope?
    \item[] Answer: \answerYes{} 
    \item[] Justification: NA
    \item[] Guidelines:
    \begin{itemize}
        \item The answer NA means that the abstract and introduction do not include the claims made in the paper.
        \item The abstract and/or introduction should clearly state the claims made, including the contributions made in the paper and important assumptions and limitations. A No or NA answer to this question will not be perceived well by the reviewers. 
        \item The claims made should match theoretical and experimental results, and reflect how much the results can be expected to generalize to other settings. 
        \item It is fine to include aspirational goals as motivation as long as it is clear that these goals are not attained by the paper. 
    \end{itemize}

\item {\bf Limitations}
    \item[] Question: Does the paper discuss the limitations of the work performed by the authors?
    \item[] Answer: \answerYes{} 
    \item[] Justification: NA
    \item[] Guidelines:
    \begin{itemize}
        \item The answer NA means that the paper has no limitation while the answer No means that the paper has limitations, but those are not discussed in the paper. 
        \item The authors are encouraged to create a separate "Limitations" section in their paper.
        \item The paper should point out any strong assumptions and how robust the results are to violations of these assumptions (e.g., independence assumptions, noiseless settings, model well-specification, asymptotic approximations only holding locally). The authors should reflect on how these assumptions might be violated in practice and what the implications would be.
        \item The authors should reflect on the scope of the claims made, e.g., if the approach was only tested on a few datasets or with a few runs. In general, empirical results often depend on implicit assumptions, which should be articulated.
        \item The authors should reflect on the factors that influence the performance of the approach. For example, a facial recognition algorithm may perform poorly when image resolution is low or images are taken in low lighting. Or a speech-to-text system might not be used reliably to provide closed captions for online lectures because it fails to handle technical jargon.
        \item The authors should discuss the computational efficiency of the proposed algorithms and how they scale with dataset size.
        \item If applicable, the authors should discuss possible limitations of their approach to address problems of privacy and fairness.
        \item While the authors might fear that complete honesty about limitations might be used by reviewers as grounds for rejection, a worse outcome might be that reviewers discover limitations that aren't acknowledged in the paper. The authors should use their best judgment and recognize that individual actions in favor of transparency play an important role in developing norms that preserve the integrity of the community. Reviewers will be specifically instructed to not penalize honesty concerning limitations.
    \end{itemize}

\item {\bf Theory assumptions and proofs}
    \item[] Question: For each theoretical result, does the paper provide the full set of assumptions and a complete (and correct) proof?
    \item[] Answer: \answerNA{} 
    \item[] Justification: NA
    \item[] Guidelines:
    \begin{itemize}
        \item The answer NA means that the paper does not include theoretical results. 
        \item All the theorems, formulas, and proofs in the paper should be numbered and cross-referenced.
        \item All assumptions should be clearly stated or referenced in the statement of any theorems.
        \item The proofs can either appear in the main paper or the supplemental material, but if they appear in the supplemental material, the authors are encouraged to provide a short proof sketch to provide intuition. 
        \item Inversely, any informal proof provided in the core of the paper should be complemented by formal proofs provided in appendix or supplemental material.
        \item Theorems and Lemmas that the proof relies upon should be properly referenced. 
    \end{itemize}

    \item {\bf Experimental result reproducibility}
    \item[] Question: Does the paper fully disclose all the information needed to reproduce the main experimental results of the paper to the extent that it affects the main claims and/or conclusions of the paper (regardless of whether the code and data are provided or not)?
    \item[] Answer: \answerYes{} 
    \item[] Justification: NA
    \item[] Guidelines:
    \begin{itemize}
        \item The answer NA means that the paper does not include experiments.
        \item If the paper includes experiments, a No answer to this question will not be perceived well by the reviewers: Making the paper reproducible is important, regardless of whether the code and data are provided or not.
        \item If the contribution is a dataset and/or model, the authors should describe the steps taken to make their results reproducible or verifiable. 
        \item Depending on the contribution, reproducibility can be accomplished in various ways. For example, if the contribution is a novel architecture, describing the architecture fully might suffice, or if the contribution is a specific model and empirical evaluation, it may be necessary to either make it possible for others to replicate the model with the same dataset, or provide access to the model. In general. releasing code and data is often one good way to accomplish this, but reproducibility can also be provided via detailed instructions for how to replicate the results, access to a hosted model (e.g., in the case of a large language model), releasing of a model checkpoint, or other means that are appropriate to the research performed.
        \item While NeurIPS does not require releasing code, the conference does require all submissions to provide some reasonable avenue for reproducibility, which may depend on the nature of the contribution. For example
        \begin{enumerate}
            \item If the contribution is primarily a new algorithm, the paper should make it clear how to reproduce that algorithm.
            \item If the contribution is primarily a new model architecture, the paper should describe the architecture clearly and fully.
            \item If the contribution is a new model (e.g., a large language model), then there should either be a way to access this model for reproducing the results or a way to reproduce the model (e.g., with an open-source dataset or instructions for how to construct the dataset).
            \item We recognize that reproducibility may be tricky in some cases, in which case authors are welcome to describe the particular way they provide for reproducibility. In the case of closed-source models, it may be that access to the model is limited in some way (e.g., to registered users), but it should be possible for other researchers to have some path to reproducing or verifying the results.
        \end{enumerate}
    \end{itemize}

\item {\bf Open access to data and code}
    \item[] Question: Does the paper provide open access to the data and code, with sufficient instructions to faithfully reproduce the main experimental results, as described in supplemental material?
    \item[] Answer: \answerYes{} 
    \item[] Justification: NA
    \item[] Guidelines:
    \begin{itemize}
        \item The answer NA means that paper does not include experiments requiring code.
        \item Please see the NeurIPS code and data submission guidelines (\url{https://nips.cc/public/guides/CodeSubmissionPolicy}) for more details.
        \item While we encourage the release of code and data, we understand that this might not be possible, so “No” is an acceptable answer. Papers cannot be rejected simply for not including code, unless this is central to the contribution (e.g., for a new open-source benchmark).
        \item The instructions should contain the exact command and environment needed to run to reproduce the results. See the NeurIPS code and data submission guidelines (\url{https://nips.cc/public/guides/CodeSubmissionPolicy}) for more details.
        \item The authors should provide instructions on data access and preparation, including how to access the raw data, preprocessed data, intermediate data, and generated data, etc.
        \item The authors should provide scripts to reproduce all experimental results for the new proposed method and baselines. If only a subset of experiments are reproducible, they should state which ones are omitted from the script and why.
        \item At submission time, to preserve anonymity, the authors should release anonymized versions (if applicable).
        \item Providing as much information as possible in supplemental material (appended to the paper) is recommended, but including URLs to data and code is permitted.
    \end{itemize}

\item {\bf Experimental setting/details}
    \item[] Question: Does the paper specify all the training and test details (e.g., data splits, hyperparameters, how they were chosen, type of optimizer, etc.) necessary to understand the results?
    \item[] Answer: \answerYes{} 
    \item[] Justification: NA
    \item[] Guidelines:
    \begin{itemize}
        \item The answer NA means that the paper does not include experiments.
        \item The experimental setting should be presented in the core of the paper to a level of detail that is necessary to appreciate the results and make sense of them.
        \item The full details can be provided either with the code, in appendix, or as supplemental material.
    \end{itemize}

\item {\bf Experiment statistical significance}
    \item[] Question: Does the paper report error bars suitably and correctly defined or other appropriate information about the statistical significance of the experiments?
    \item[] Answer: \answerNo{} 
    \item[] Justification: NA
    \item[] Guidelines:
    \begin{itemize}
        \item The answer NA means that the paper does not include experiments.
        \item The authors should answer "Yes" if the results are accompanied by error bars, confidence intervals, or statistical significance tests, at least for the experiments that support the main claims of the paper.
        \item The factors of variability that the error bars are capturing should be clearly stated (for example, train/test split, initialization, random drawing of some parameter, or overall run with given experimental conditions).
        \item The method for calculating the error bars should be explained (closed form formula, call to a library function, bootstrap, etc.)
        \item The assumptions made should be given (e.g., Normally distributed errors).
        \item It should be clear whether the error bar is the standard deviation or the standard error of the mean.
        \item It is OK to report 1-sigma error bars, but one should state it. The authors should preferably report a 2-sigma error bar than state that they have a 96\% CI, if the hypothesis of Normality of errors is not verified.
        \item For asymmetric distributions, the authors should be careful not to show in tables or figures symmetric error bars that would yield results that are out of range (e.g. negative error rates).
        \item If error bars are reported in tables or plots, The authors should explain in the text how they were calculated and reference the corresponding figures or tables in the text.
    \end{itemize}

\item {\bf Experiments compute resources}
    \item[] Question: For each experiment, does the paper provide sufficient information on the computer resources (type of compute workers, memory, time of execution) needed to reproduce the experiments?
    \item[] Answer: \answerYes{} 
    \item[] Justification: NA
    \item[] Guidelines:
    \begin{itemize}
        \item The answer NA means that the paper does not include experiments.
        \item The paper should indicate the type of compute workers CPU or GPU, internal cluster, or cloud provider, including relevant memory and storage.
        \item The paper should provide the amount of compute required for each of the individual experimental runs as well as estimate the total compute. 
        \item The paper should disclose whether the full research project required more compute than the experiments reported in the paper (e.g., preliminary or failed experiments that didn't make it into the paper). 
    \end{itemize}
    
\item {\bf Code of ethics}
    \item[] Question: Does the research conducted in the paper conform, in every respect, with the NeurIPS Code of Ethics \url{https://neurips.cc/public/EthicsGuidelines}?
    \item[] Answer: \answerYes{} 
    \item[] Justification: NA
    \item[] Guidelines:
    \begin{itemize}
        \item The answer NA means that the authors have not reviewed the NeurIPS Code of Ethics.
        \item If the authors answer No, they should explain the special circumstances that require a deviation from the Code of Ethics.
        \item The authors should make sure to preserve anonymity (e.g., if there is a special consideration due to laws or regulations in their jurisdiction).
    \end{itemize}

\item {\bf Broader impacts}
    \item[] Question: Does the paper discuss both potential positive societal impacts and negative societal impacts of the work performed?
    \item[] Answer: \answerNA{} 
    \item[] Justification: NA
    \item[] Guidelines:
    \begin{itemize}
        \item The answer NA means that there is no societal impact of the work performed.
        \item If the authors answer NA or No, they should explain why their work has no societal impact or why the paper does not address societal impact.
        \item Examples of negative societal impacts include potential malicious or unintended uses (e.g., disinformation, generating fake profiles, surveillance), fairness considerations (e.g., deployment of technologies that could make decisions that unfairly impact specific groups), privacy considerations, and security considerations.
        \item The conference expects that many papers will be foundational research and not tied to particular applications, let alone deployments. However, if there is a direct path to any negative applications, the authors should point it out. For example, it is legitimate to point out that an improvement in the quality of generative models could be used to generate deepfakes for disinformation. On the other hand, it is not needed to point out that a generic algorithm for optimizing neural networks could enable people to train models that generate Deepfakes faster.
        \item The authors should consider possible harms that could arise when the technology is being used as intended and functioning correctly, harms that could arise when the technology is being used as intended but gives incorrect results, and harms following from (intentional or unintentional) misuse of the technology.
        \item If there are negative societal impacts, the authors could also discuss possible mitigation strategies (e.g., gated release of models, providing defenses in addition to attacks, mechanisms for monitoring misuse, mechanisms to monitor how a system learns from feedback over time, improving the efficiency and accessibility of ML).
    \end{itemize}
    
\item {\bf Safeguards}
    \item[] Question: Does the paper describe safeguards that have been put in place for responsible release of data or models that have a high risk for misuse (e.g., pretrained language models, image generators, or scraped datasets)?
    \item[] Answer: \answerNA{} 
    \item[] Justification: NA
    \item[] Guidelines:
    \begin{itemize}
        \item The answer NA means that the paper poses no such risks.
        \item Released models that have a high risk for misuse or dual-use should be released with necessary safeguards to allow for controlled use of the model, for example by requiring that users adhere to usage guidelines or restrictions to access the model or implementing safety filters. 
        \item Datasets that have been scraped from the Internet could pose safety risks. The authors should describe how they avoided releasing unsafe images.
        \item We recognize that providing effective safeguards is challenging, and many papers do not require this, but we encourage authors to take this into account and make a best faith effort.
    \end{itemize}

\item {\bf Licenses for existing assets}
    \item[] Question: Are the creators or original owners of assets (e.g., code, data, models), used in the paper, properly credited and are the license and terms of use explicitly mentioned and properly respected?
    \item[] Answer: \answerYes{} 
    \item[] Justification: NA
    \item[] Guidelines:
    \begin{itemize}
        \item The answer NA means that the paper does not use existing assets.
        \item The authors should cite the original paper that produced the code package or dataset.
        \item The authors should state which version of the asset is used and, if possible, include a URL.
        \item The name of the license (e.g., CC-BY 4.0) should be included for each asset.
        \item For scraped data from a particular source (e.g., website), the copyright and terms of service of that source should be provided.
        \item If assets are released, the license, copyright information, and terms of use in the package should be provided. For popular datasets, \url{paperswithcode.com/datasets} has curated licenses for some datasets. Their licensing guide can help determine the license of a dataset.
        \item For existing datasets that are re-packaged, both the original license and the license of the derived asset (if it has changed) should be provided.
        \item If this information is not available online, the authors are encouraged to reach out to the asset's creators.
    \end{itemize}

\item {\bf New assets}
    \item[] Question: Are new assets introduced in the paper well documented and is the documentation provided alongside the assets?
    \item[] Answer: \answerYes{} 
    \item[] Justification: NA
    \item[] Guidelines:
    \begin{itemize}
        \item The answer NA means that the paper does not release new assets.
        \item Researchers should communicate the details of the dataset/code/model as part of their submissions via structured templates. This includes details about training, license, limitations, etc. 
        \item The paper should discuss whether and how consent was obtained from people whose asset is used.
        \item At submission time, remember to anonymize your assets (if applicable). You can either create an anonymized URL or include an anonymized zip file.
    \end{itemize}

\item {\bf Crowdsourcing and research with human subjects}
    \item[] Question: For crowdsourcing experiments and research with human subjects, does the paper include the full text of instructions given to participants and screenshots, if applicable, as well as details about compensation (if any)? 
    \item[] Answer: \answerNA{} 
    \item[] Justification: NA
    \item[] Guidelines:
    \begin{itemize}
        \item The answer NA means that the paper does not involve crowdsourcing nor research with human subjects.
        \item Including this information in the supplemental material is fine, but if the main contribution of the paper involves human subjects, then as much detail as possible should be included in the main paper. 
        \item According to the NeurIPS Code of Ethics, workers involved in data collection, curation, or other labor should be paid at least the minimum wage in the country of the data collector. 
    \end{itemize}

\item {\bf Institutional review board (IRB) approvals or equivalent for research with human subjects}
    \item[] Question: Does the paper describe potential risks incurred by study participants, whether such risks were disclosed to the subjects, and whether Institutional Review Board (IRB) approvals (or an equivalent approval/review based on the requirements of your country or institution) were obtained?
    \item[] Answer: \answerNA{} 
    \item[] Justification: NA
    \item[] Guidelines:
    \begin{itemize}
        \item The answer NA means that the paper does not involve crowdsourcing nor research with human subjects.
        \item Depending on the country in which research is conducted, IRB approval (or equivalent) may be required for any human subjects research. If you obtained IRB approval, you should clearly state this in the paper. 
        \item We recognize that the procedures for this may vary significantly between institutions and locations, and we expect authors to adhere to the NeurIPS Code of Ethics and the guidelines for their institution. 
        \item For initial submissions, do not include any information that would break anonymity (if applicable), such as the institution conducting the review.
    \end{itemize}

\item {\bf Declaration of LLM usage}
    \item[] Question: Does the paper describe the usage of LLMs if it is an important, original, or non-standard component of the core methods in this research? Note that if the LLM is used only for writing, editing, or formatting purposes and does not impact the core methodology, scientific rigorousness, or originality of the research, declaration is not required.
    \item[] Answer: \answerNA{} 
    \item[] Justification: NA
    \item[] Guidelines:
    \begin{itemize}
        \item The answer NA means that the core method development in this research does not involve LLMs as any important, original, or non-standard components.
        \item Please refer to our LLM policy (\url{https://neurips.cc/Conferences/2025/LLM}) for what should or should not be described.
    \end{itemize}

\end{enumerate}

\clearpage

\setcounter{page}{1}
\setcounter{section}{0}

\renewcommand\thesection{\Alph{section}}
\section{Introduction}
\label{sec_supp_intro}
We first provide additional details of our method, including parameter settings and further descriptions, in Sec.~\ref{sec_add_detail}. Then, we present extended experimental results in Sec.~\ref{sec_add_exp}. In Sec.~\ref{sec_add_discussion}, we conduct deeper analyses on entropy in relation to model behavior and image content, along with more visualizations of entropy maps. Sec.~\ref{sec_supp_conclusion} discusses potential limitations and future directions. Lastly, we include more visual comparisons of the proposed method in Sec.~\ref{sec_add_visual_comparison}.
\section{Additional details of method}
\label{sec_add_detail}

\subsection{Hyperparameter settings in Sec. 3.2 in maintext}
In Sec. 3.2 in maintext, we propose to dynamically control the sampling temperature based on entropy. However, due to significant differences between base models, it is difficult to apply the same parameters across all settings. Therefore, we list the detailed parameters for each model in Table~\ref{tab_supp_hyper_param_setting}. For undertrained models such as LlamaGen stage1, higher randomness is required at low-entropy stages to avoid generating large areas of repetitive tokens. In contrast, well-trained models benefit from a smoother temperature schedule.

\begin{table}[h]
\centering
\setlength{\abovecaptionskip}{0.1cm}
\setlength{\belowcaptionskip}{0.1cm}
\caption{Hyperparameter settings of different models.}
\setlength{\tabcolsep}{4mm}{
\resizebox{0.4\columnwidth}{!}{%
\begin{tabular}{@{}l|ccc@{}}
\toprule
            & $T_0$ & $\alpha$ & $\theta$ \\ \midrule
LlamaGen    & 2.5 & 3.0  & 0.6  \\
Lumina-mGPT & 2.0 & 2.5  & 0.6   \\
Meissonic   & 2.5 & 3.0  & 0.7   \\
STAR        & 2.5 & 3.0  & 0.5   \\ \bottomrule
\end{tabular}
}}
\label{tab_supp_hyper_param_setting}
\vspace{-3mm}
\end{table}

\subsection{Detailed description of speculative decoding in images}
We accelerate inference based on existing speculative decoding schemes~\cite{teng2024sjd} in Sec. 3.4 in maintext, thereby further reducing inference cost without sacrificing output quality. Due to space constraints, we did not elaborate on the baseline speculative decoding methods in the main text. Here, we provide more details.

This method aims to accelerate auto-regressive text-to-image generation by allowing multiple tokens to be generated in parallel without training. Inspired by speculative decoding, SJD introduces a probabilistic acceptance criterion that compares the confidence of draft tokens from two consecutive iterations. In each iteration $j$, given a draft token $x_i^{(j)}$, SJD computes its acceptance probability based on the ratio between two conditional probabilities:
\begin{equation}
r < \min\left(1, \frac{p_{\theta}(x_i^{(j)} \mid x_{1:i-1}^{(j)})}{p_{\theta}(x_i^{(j)} \mid x_{1:i-1}^{(j-1)})}\right),
\end{equation}
where $r \sim \mathcal{U}[0,1]$. Accepted tokens are fixed, while the others are resampled from a calibrated distribution:
\begin{equation}
x_i^{(j+1)} \sim \frac{\max(0, p_{\theta}(x \mid x_{1:i-1}^{(j)}) - p_{\theta}(x \mid x_{1:i-1}^{(j-1)}))}{\sum_x \max(0, \cdot)}.
\end{equation}
This allows high-randomness sampling, crucial for image diversity, while significantly reducing decoding steps. SJD operates in a windowed, iterative manner and supports optional spatially-informed token initialization to further improve efficiency.

\section{Additional experimental results}
\label{sec_add_exp}

\subsection{Generation performance on an additional dataset}
Since the COCO2017 dataset used in the main experiments contains only 5,000 images, it may lead to slight estimation bias in the FID computation, as FID becomes more reliable with larger sample sizes. To assess the potential misjudgment of model performance caused by limited image numbers, we further evaluated the metrics on a larger dataset, COCO2014, as shown in Table~\ref{tab_supp_coco2014}.

\begin{table}[]
\centering
\caption{Evaluation on the larger COCO2014 dataset (compared to COCO2017 in the main text). 
The results demonstrate that the improvements brought by our method remain consistent and significant across datasets.}
\setlength{\tabcolsep}{5.5mm}{
\resizebox{0.7\columnwidth}{!}{%
\begin{tabular}{@{}llc|cc@{}}
\toprule
\textbf{Model} & \textbf{Method} &  & \textbf{FID}$\downarrow$ & \textbf{CLIP-Score}$\uparrow$ \\ 
\midrule
\multirow{2}{*}{LlamaGen~\cite{sun2024llamagen}} 
 & Baseline &  & 13.37 & \textbf{0.2561} \\
 & Ours     &  & \textbf{11.59} & 0.2560 \\ 
\midrule
\multirow{2}{*}{Meissonic~\cite{bai2024meissonic}} 
 & Baseline &  & 44.54 & 0.2567 \\
 & Ours+Mask &  & \textbf{38.95} & \textbf{0.2590} \\ 
\midrule
\multirow{2}{*}{STAR~\cite{ma2024star}} 
 & Baseline &  & 24.86 & 0.2581 \\
 & Ours+Scale &  & \textbf{22.09} & \textbf{0.2598} \\ 
\midrule
\multirow{2}{*}{Lumina-mGPT~\cite{liu2024lumina_mgpt}} 
 & Baseline &  & 18.23 & 0.2641 \\
 & Ours &  & \textbf{16.16} & \textbf{0.2659} \\ 
\bottomrule
\end{tabular}
}}
\label{tab_supp_coco2014}
\vspace{-3mm}
\end{table}

\subsection{Effect of random seeds on model performance}

Due to the autoregressive nature of our model, each token is sampled from a probability distribution, making the generated images sensitive to random seeds. Specifically, we observed that metrics such as FID, CLIP-Score, DPG, and HPS may vary with different random seeds.
To further analyze this effect, we randomly selected 10 seeds from the range $[0, 1e6]$, ran the generation model 10 times under these conditions, and computed the mean and standard deviation of the results. As shown in the table below, random seeds have little impact on the performance gain introduced by our method, further confirming the robustness and effectiveness of our approach. See Table~\ref{tab_supp_rand_seed}.

\begin{table}[h]
\centering
\setlength{\abovecaptionskip}{0.1cm}
\setlength{\belowcaptionskip}{0.1cm}
\caption{Mean and standard deviation over 10 random seeds. Our method consistently outperforms the baseline with statistically significant improvements.}
\setlength{\tabcolsep}{2mm}{
\resizebox{1.0\columnwidth}{!}{%
\begin{tabular}{l|cc|cc|cc}
\toprule
     & \multicolumn{2}{c|}{LlamaGen}        & \multicolumn{2}{c|}{Meissonic} & \multicolumn{2}{c}{STAR} \\ \cline{2-7} 
     & Baseline                     & Ours & Baaseline        & Ours       & Baseline      & Ours     \\ \hline
FID$\downarrow$  & 21.79 $\pm$ 0.16 & 20.24 $\pm$ 0.08      & 53.31 $\pm$ 0.33 & 48.18 $\pm$ 0.29            & 35.48 $\pm$ 0.29 & 33.12 $\pm$ 0.49          \\
CLIP$\uparrow$ & 25.95 $\pm$ 0.02 & 25.95 $\pm$ 0.03      & 25.30 $\pm$ 0.02 & 25.62 $\pm$ 0.03            & 25.47 $\pm$ 0.02 & 25.63 $\pm$ 0.03          \\
DPG$\uparrow$  & 43.74 $\pm$ 0.34 & 48.87 $\pm$ 0.32      & 64.07 $\pm$ 0.13 & 66.91 $\pm$ 0.17          & 70.28 $\pm$ 0.12 & 70.40 $\pm$ 0.11          \\
HPS$\uparrow$  & 21.23 $\pm$ 0.03 & 21.40 $\pm$ 0.04      & 29.33 $\pm$ 0.03 & 30.06 $\pm$ 0.02            & 28.70 $\pm$ 0.08 & 29.07 $\pm$ 0.11          \\ 
\bottomrule
\end{tabular}
}}
\label{tab_supp_rand_seed}
\vspace{-3mm}
\end{table}

\subsection{Entropy $\&$ top-$p$ and temperature}
\begin{figure*}[h]
\setlength{\abovecaptionskip}{0.1cm}
\setlength{\belowcaptionskip}{0.1cm}
\begin{center}
\includegraphics[width=0.6\textwidth]{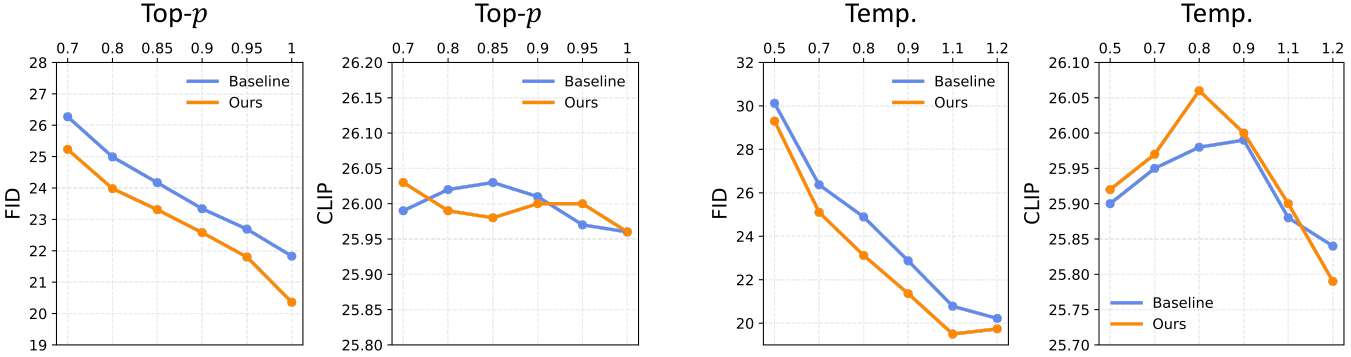}
\end{center}
\caption{
Combination of our sampling strategy with existing methods (Top-$p$, temperature). ``Temp.'' is short for temperature.
}
\vspace{-0.3cm}
\label{fig_supp_ab_param_2}
\end{figure*}

In the main text, we analyze the relationship between our entropy-based sampling strategy and existing sampling parameters such as CFG and top-$K$. By combining our method with these parameters, we observe improved robustness, reducing sensitivity to hyperparameter choices and yielding better FID and CLIP-Score. Here, we further examine other sampling parameters—top-p and temperature—which are rarely used in autoregressive models due to their tendency to distort the output distribution and severely degrade either FID or CLIP-Score. Comparative results between our method and the baseline are shown in Fig.~\ref{fig_supp_ab_param_2}.

\subsection{Additional comparison with top-$K$ and CFG}
\begin{figure*}[h]
\setlength{\abovecaptionskip}{0.1cm}
\setlength{\belowcaptionskip}{0.1cm}
\begin{center}
\includegraphics[width=0.6\textwidth]{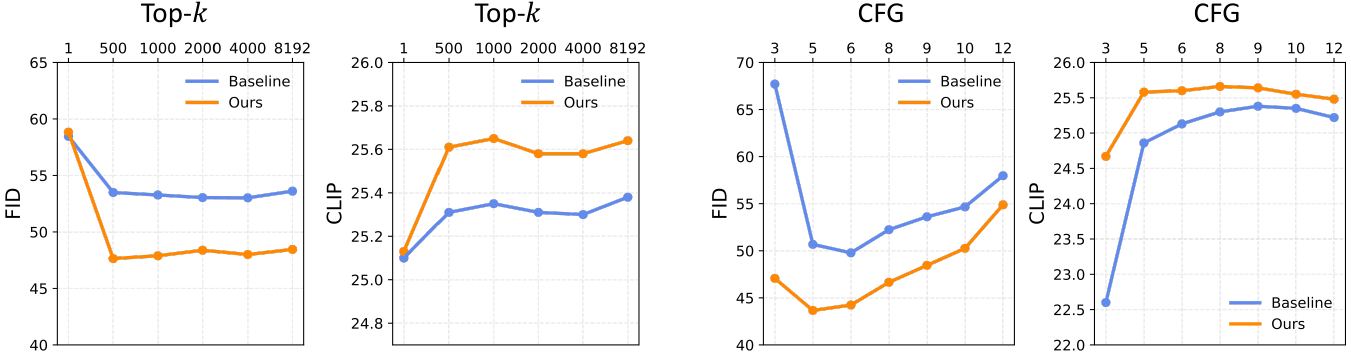}
\end{center}
\caption{
Combination of our sampling strategy with existing methods (Top-$k$, CFG) on Meissonic.
}
\vspace{-0.3cm}
\label{fig_supp_ab_param_meissonic}
\end{figure*}

\begin{figure*}[h]
\setlength{\abovecaptionskip}{0.1cm}
\setlength{\belowcaptionskip}{0.1cm}
\begin{center}
\includegraphics[width=0.6\textwidth]{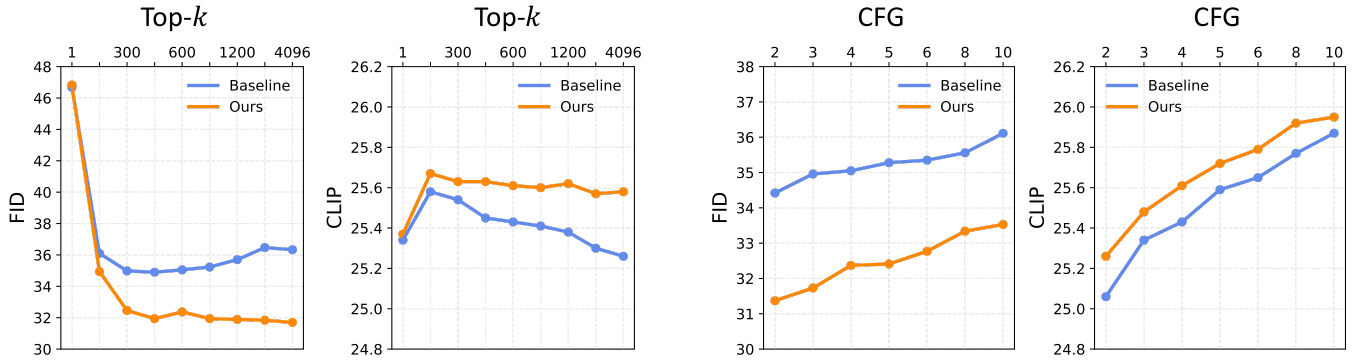}
\end{center}
\caption{
Combination of our sampling strategy with existing methods (Top-$k$, CFG) on STAR.
}
\vspace{-0.3cm}
\label{fig_supp_ab_param_star}
\end{figure*}

In Sec.~4.4 of the main paper, we discussed the differences between our method and existing sampling strategies (Top-$K$ and CFG) using LlamaGen. Here, we provide additional comparisons on other models to analyze the relationship between entropy-aware temperature and these conventional sampling approaches, results are presented in Fig.~\ref{fig_supp_ab_param_meissonic} and Fig.~\ref{fig_supp_ab_param_star}.
Consistent with our observations in Sec. 4.4 in maintext, the proposed strategy mitigates performance fluctuations caused by hyperparameter choices (e.g., CFG and top-K), leading to a better balance between fidelity and text-image alignment.


\subsection{Additional evaluation of performance regarding temperature}
\begin{figure*}[h]
\setlength{\abovecaptionskip}{0.1cm}
\setlength{\belowcaptionskip}{0.1cm}
\begin{center}
\includegraphics[width=0.8\textwidth]{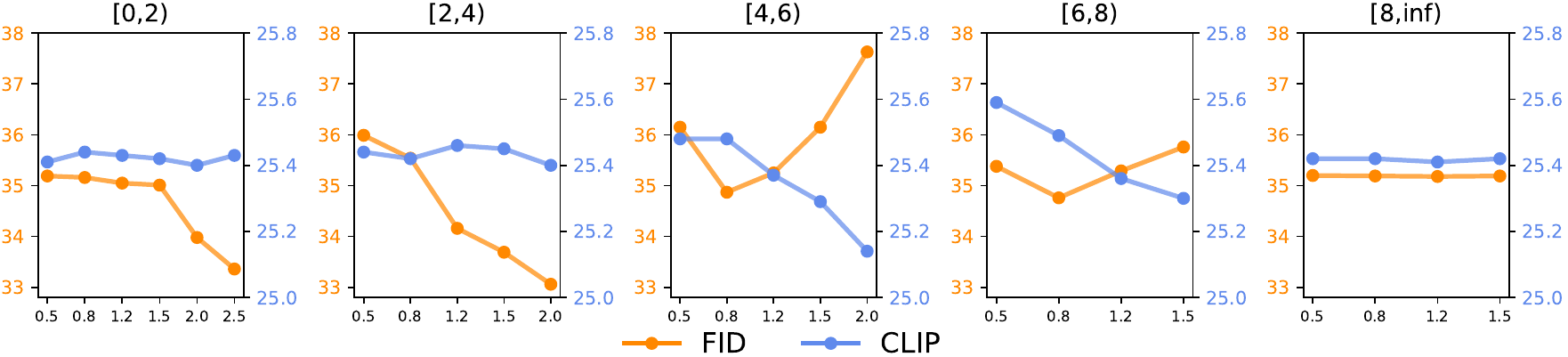}
\end{center}
\caption{
Varying temperature by entropy range affects FID and CLIP score: lower-entropy tokens benefit from higher temperatures, and vice versa. Experiments are conducted on STAR using the COCO2017 validation split by varying temperature across entropy ranges.
}
\vspace{-0.3cm}
\label{fig_supp_ab_temperature_star}
\end{figure*}

In the main text, we analyzed how adjusting the sampling temperature of tokens in different entropy ranges affects the generation quality for LlamaGen. Here, we further extend the study to more models. See Fig.~\ref{fig_supp_ab_temperature_star}.

\section{Additional discussion about entropy}
\label{sec_add_discussion}

\subsection{Entropy $\&$ generative models}
\subsubsection{Visualization of entropy \& images}
\begin{figure}[h]
\setlength{\abovecaptionskip}{0.2cm}
\setlength{\belowcaptionskip}{0.2cm}
  \centering
   \includegraphics[width=0.8\linewidth]{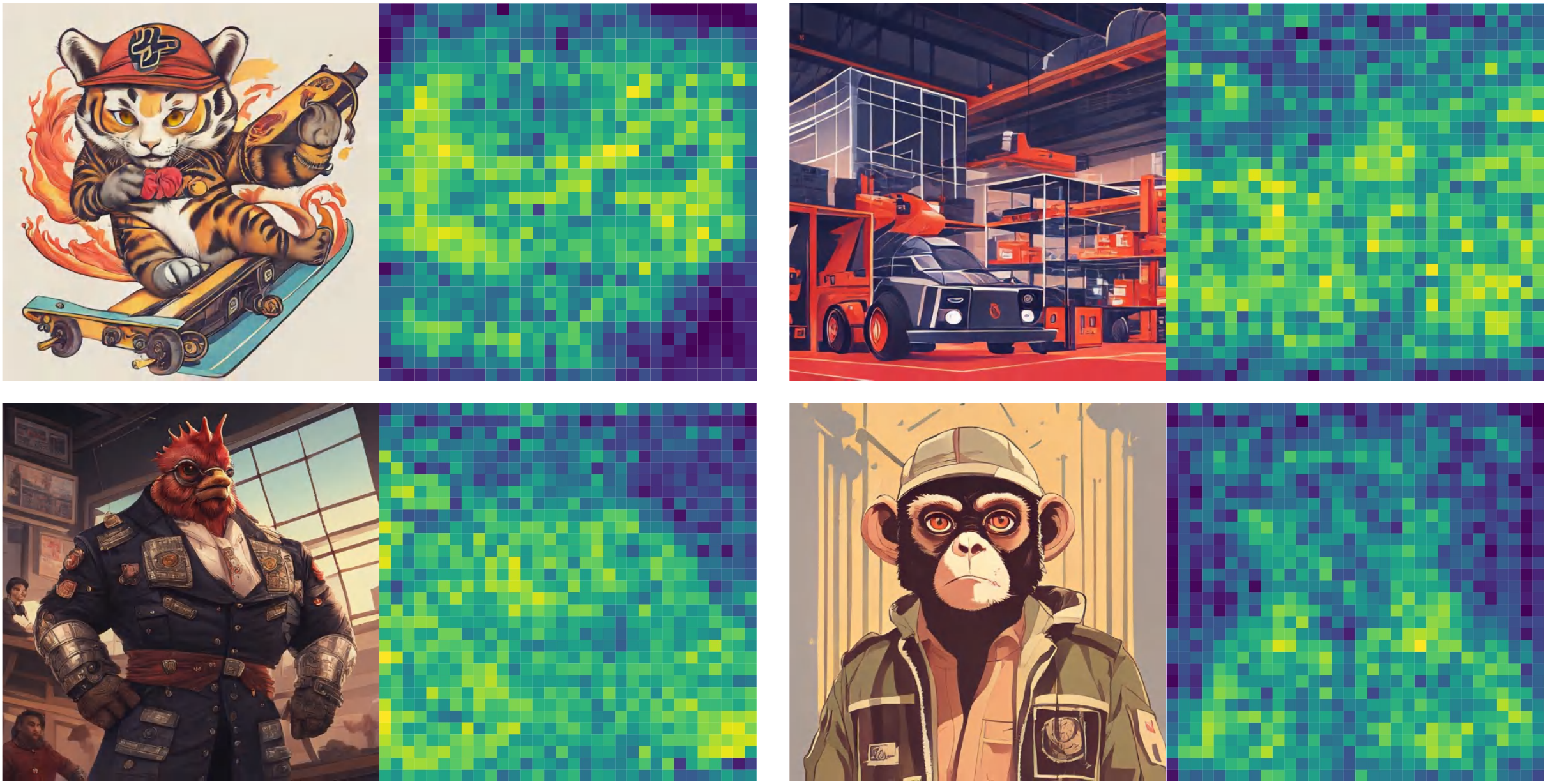}

   \caption{Entropy visualization of LlamaGen.}
   \label{fig_entropy_view_llamagen}
\end{figure}

\begin{figure}[h]
\setlength{\abovecaptionskip}{0.2cm}
\setlength{\belowcaptionskip}{0.2cm}
  \centering
   \includegraphics[width=0.8\linewidth]{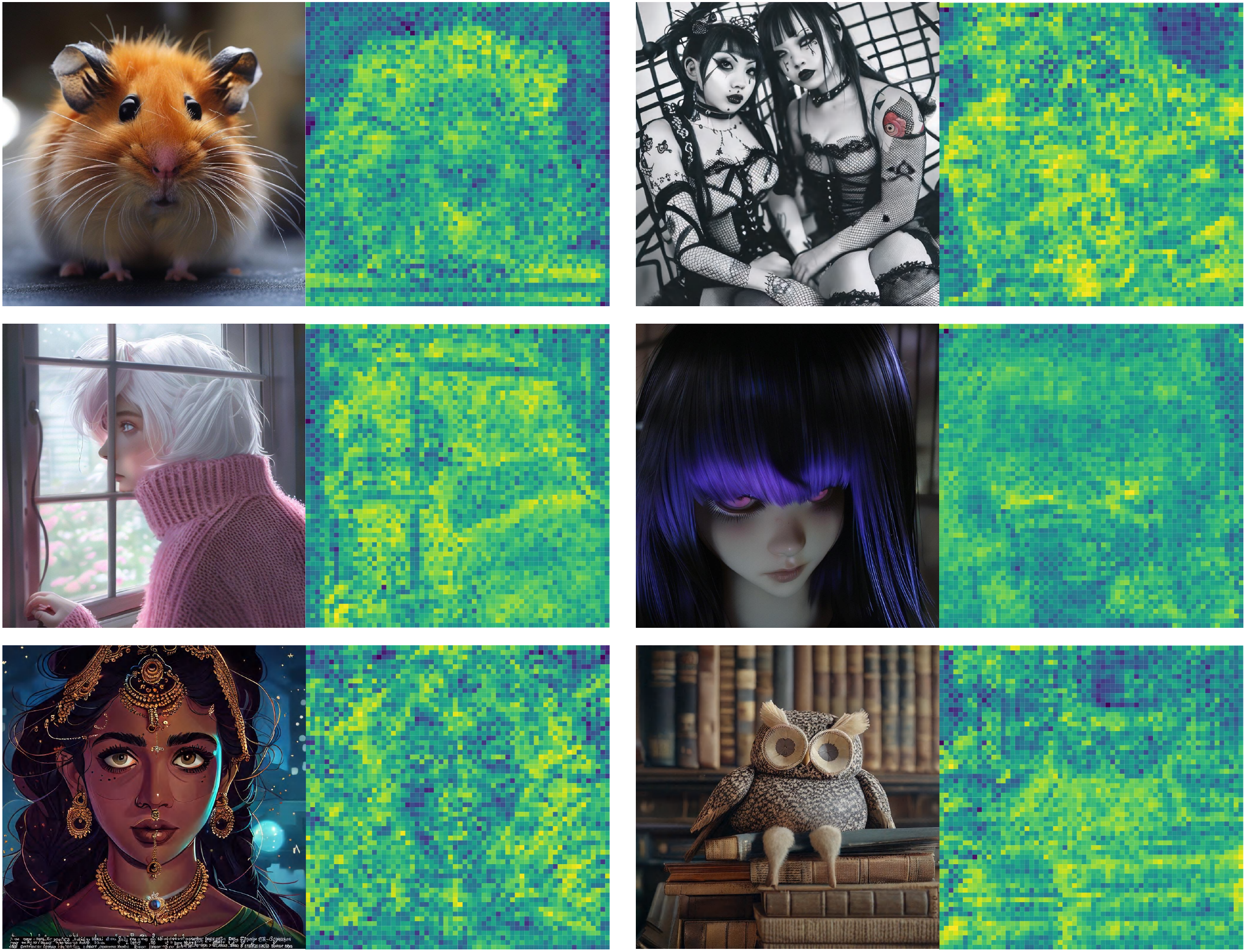}

   \caption{Entropy visualization of Lumina-mGPT.}
   \label{fig_entropy_view_mgpt}
\end{figure}

Due to space limitations in the main text, we did not provide extended entropy visualizations and analysis. Here, we include additional entropy maps for LlamaGen and Lumina-mGPT. See Fig.~\ref{fig_entropy_view_llamagen} and Fig.~\ref{fig_entropy_view_mgpt}.

\subsubsection{Mask-prediction models}
\begin{figure}[h]
\setlength{\abovecaptionskip}{0.2cm}
\setlength{\belowcaptionskip}{0.2cm}
  \centering
   \includegraphics[width=0.8\linewidth]{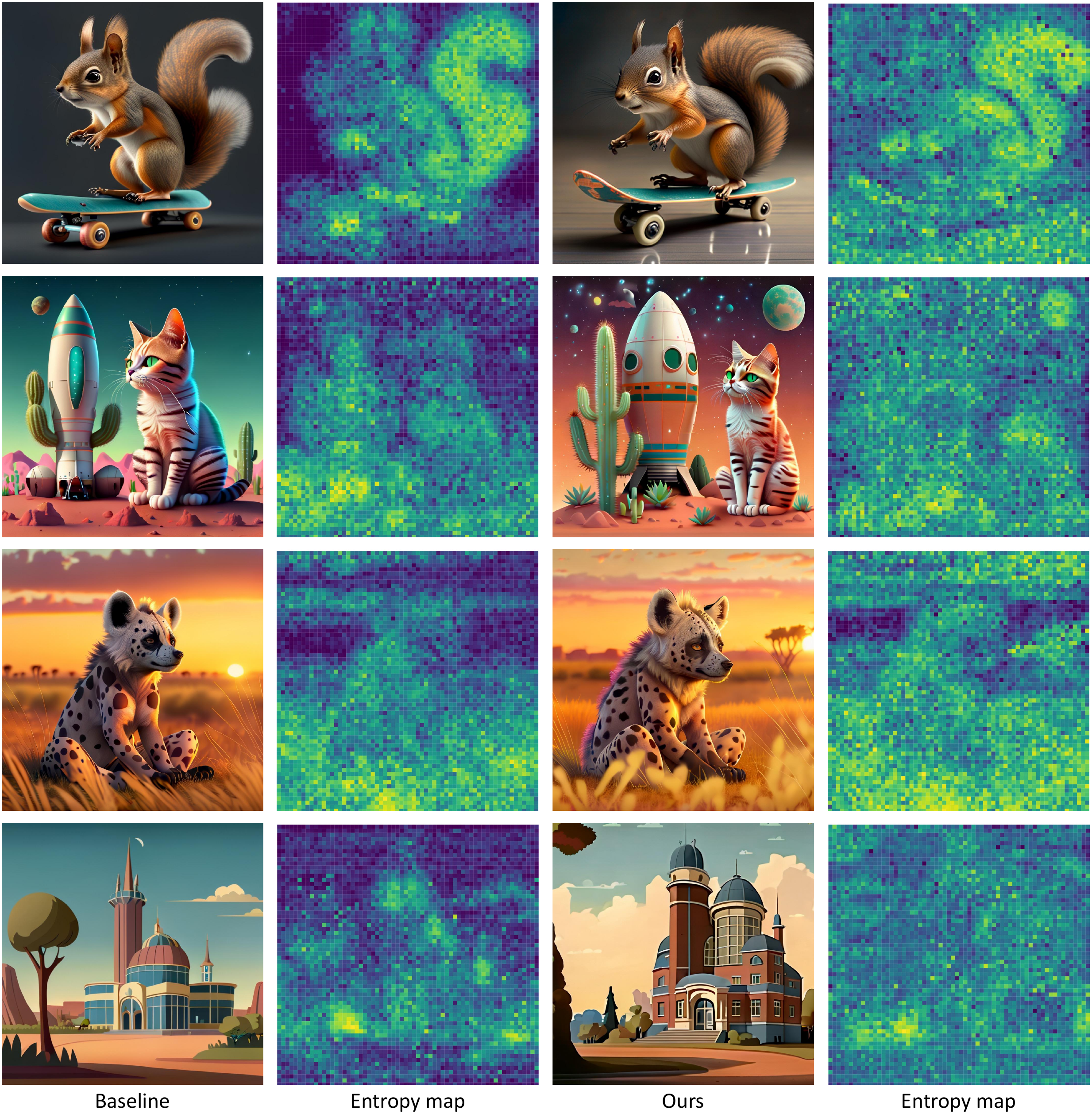}

   \caption{Entropy visualization of Meissonic. Our entropy-based temperature leads to a more spatially balanced entropy distribution and enables richer image content.}
   \label{fig_entropy_view_meissonic}
\end{figure}

We provide additional entropy-based analysis of the mask model. Since the generation involves multiple timesteps, where a subset of tokens is accepted at each step based on previously generated content, we compute the entropy of accepted tokens at each timestep and aggregate them into a final entropy map. As shown in Fig.~\ref{fig_entropy_view_meissonic}, applying the proposed entropy-based temperature leads to a more spatially balanced entropy distribution and enables richer image content while maintaining generation stability.

\begin{figure*}[h]
\setlength{\abovecaptionskip}{0.1cm}
\setlength{\belowcaptionskip}{0.1cm}
\begin{center}
\includegraphics[width=0.6\textwidth]{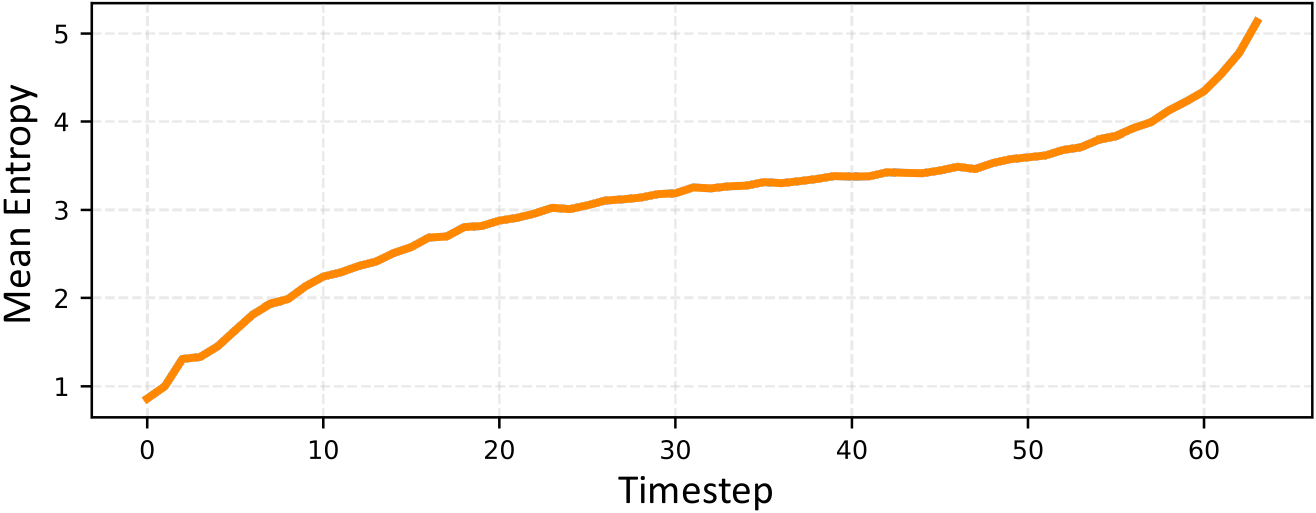}
\end{center}
\caption{
Mean entropy of each step from mask-prediction model. Values are averaged from $\sim$100 generated images.
}
\vspace{-0.3cm}
\label{fig_supp_mask_entropy_mean}
\end{figure*}

In addition, we further analyze the average entropy of tokens accepted at each timestep, as shown in Fig.~\ref{fig_supp_mask_entropy_mean}. As discussed in the main text, due to the confidence-based token selection strategy, tokens accepted in earlier steps tend to have lower entropy, since they are more likely to receive high confidence scores. In contrast, tokens accepted in later steps (\textgreater60) exhibit significantly higher entropy. Moreover, more tokens are accepted in these later stages, which increases the risk of violating the autoregressive assumption that spatially adjacent tokens should be sampled as independently as possible. This may lead to degraded image quality. Therefore, adopting a more conservative sampling strategy for these high-entropy tokens could help improve the overall generation quality.

\subsubsection{Scale-wise models}
\begin{figure}[h]
\setlength{\abovecaptionskip}{0.2cm}
\setlength{\belowcaptionskip}{0.2cm}
  \centering
   \includegraphics[width=1\linewidth]{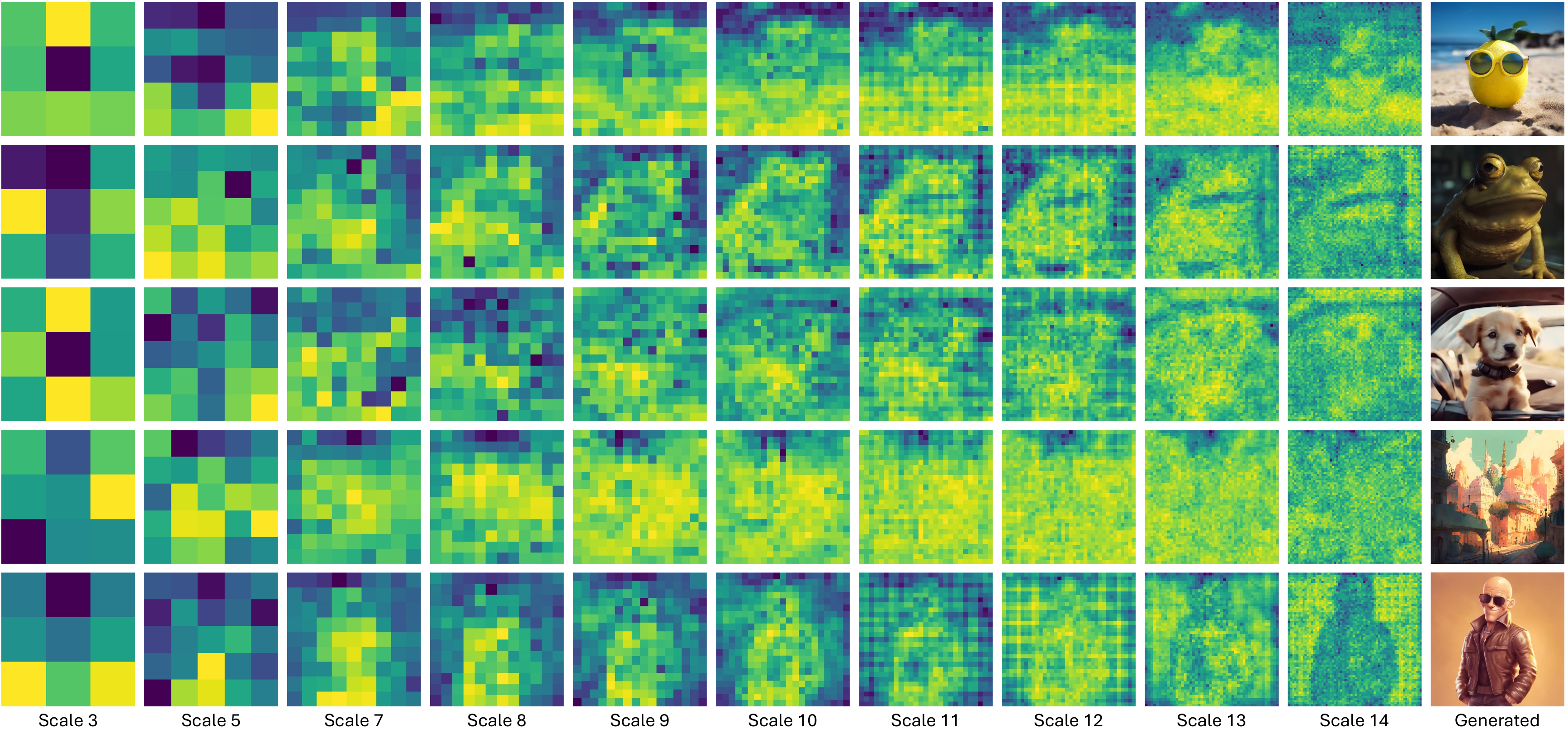}

   \caption{Entropy visualization of STAR. .}
   \label{fig_entropy_view_star}
\end{figure}

For the scale-wise model, the generation process constructs a complete image by predicting logits maps at multiple scales. Each scale is conditioned on the residuals from the preceding scales, meaning that the sum of the feature maps generated at all scales is passed through the detokenizer to form the final output. In this generation paradigm, different scales exhibit distinct roles. Specifically, as described in~\cite{wang2025varedit}, the earlier scales are responsible for generating the main structure of the image, while the later scales refine the result with fine details such as texture. We visualize the entropy maps of each scale during generation, as shown in Fig.~\ref{fig_entropy_view_star}. From scale 8 to scale 12, the model tends to focus more on the foreground, with significantly higher entropy observed in the regions corresponding to the primary subject. In contrast, at scale 13 and 14, there is no clear bias between foreground and background, indicating a more uniform attention across the image.

\begin{figure*}[h]
\setlength{\abovecaptionskip}{0.1cm}
\setlength{\belowcaptionskip}{0.1cm}
\begin{center}
\includegraphics[width=0.6\textwidth]{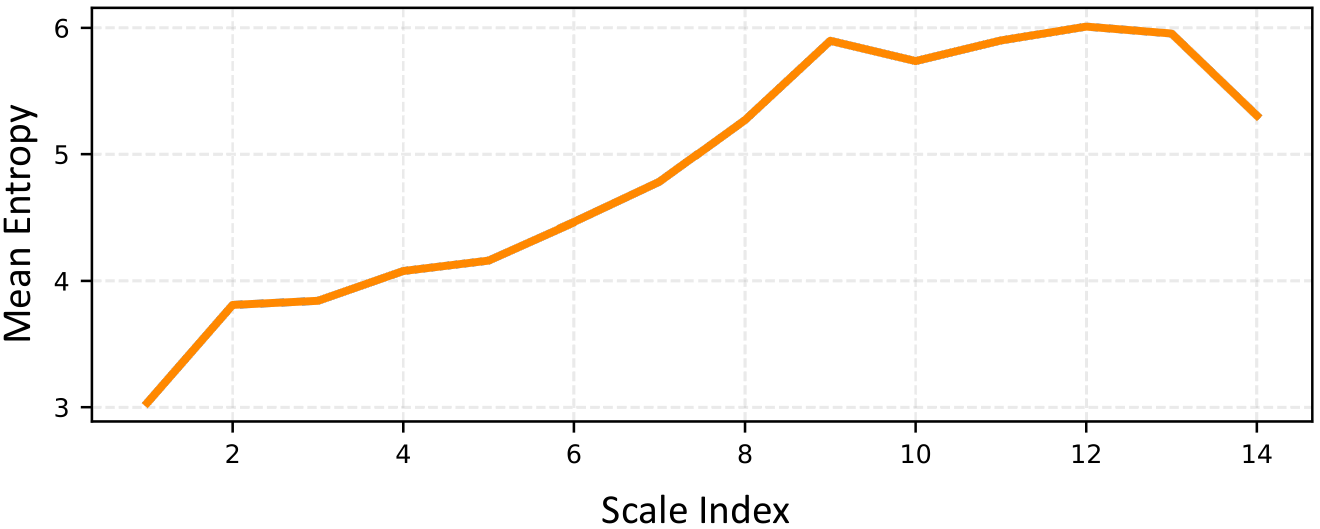}
\end{center}
\caption{
Mean entropy of each scale from scale-wise model. Values are averaged from $\sim$100 generated images.
}
\vspace{-0.3cm}
\label{fig_supp_star_entropy_mean}
\end{figure*}

In addition, we compute the average entropy for each scale, as shown in Fig.~\ref{fig_supp_star_entropy_mean}. The later scales exhibit relatively higher entropy, while the earlier scales tend to have lower average entropy (however a decreasing trend is observed in the final two scales). This further indicates that different scales carry varying amounts of information.

\subsection{Entropy $\&$ generated contents}
In practice, the logits are not simply positively correlated with the complexity of image content. We observe that regions with clear, well-defined content do not always exhibit high entropy; instead, their entropy typically falls within a moderate range (e.g., between 2 and 8). The more deterministic the content, the lower the entropy tends to be. In contrast, regions with entropy lower than 2 or higher than 8 often correspond to simple backgrounds or overly complex, unfaithful details. Especially for regions with entropy above 8, the generated details are frequently meaningless. This also explains why adjusting the logits in these low- and high-entropy areas, as discussed in our motivation experiment, does not significantly harm text-image alignment.

\subsection{Is entropy the best indicator for information?}
From a theoretical perspective, the entropy of logits reflects the model's confidence in predicting the current token. When the model is sufficiently trained—or when its capacity is strong—it may produce low entropy even in semantically important regions. In practice, we observe cases where foreground objects (e.g., faces) yield lower entropy than complex backgrounds. This suggests that the model's confidence is not solely determined by information density, but also by the number of plausible token candidates in a region. For instance, highly structured areas like faces tend to have a unique correct token and thus low uncertainty, despite containing rich semantic information. In contrast, cluttered textures such as grass or foliage may allow for more varied token predictions, resulting in higher entropy.

Based on the above analysis, entropy may need to be combined with additional indicators to more accurately characterize the information distribution within an image.
Specifically, more precise token-wise handling can be achieved by incorporating the similarity among top-ranked tokens in the logits distribution. For instance, if the entropy is low but the top tokens are not similar, the prediction can be deemed accurate; however, if the top tokens are highly similar under low entropy, the randomness at that position may need to be further increased. Conversely, under high-entropy conditions, a set of similar top tokens may indicate the existence of genuinely diverse possibilities. Moreover, analyzing the similarity of logits between adjacent tokens could help identify tokens that require more precise predictions—for example, if a token’s probability distribution significantly differs from that of the previous token, it may warrant stricter sampling, regardless of its entropy level. We leave these directions for future exploration.

Moreover, since dynamic temperature only adjusts the randomness of the probability distribution (i.e., the variance of the logits) but not the location of its peak, further combining it with CFG may help achieve better performance.
\section{Future works \& limitations}
\label{sec_supp_conclusion}
\subsection{Broader impacts}
This work is the first to explore the decoding problem in autoregressive visual generation, highlighting the inherent differences between image and text generation. While our approach may not be fully complete and still leaves room for improvement, we hope it can inspire future research to further investigate this issue and develop decoding strategies tailored specifically for visual generation, ultimately advancing unified multimodal generation.

\subsection{Future works}
Currently, we propose a training-free sampling strategy for image generation by adaptively controlling sampling randomness based on the distribution of predicted logits. However, this approach is sensitive to hyperparameters, and due to significant differences across backbone architectures, optimal settings vary across models. Moreover, as a simple inference-time method built upon pretrained models, its performance gains may be limited for certain models.

In the future, this strategy could be integrated into the training framework for further performance improvement or acceleration. For example, it may be combined with early-exit mechanisms to allocate computation dynamically across tokens, or used to guide training by leveraging entropy to focus more on informative regions, thus accelerating convergence.

\subsection{Limitations}
The proposed method mainly mitigates issues caused by inconsistent token sampling strategies under varying information densities, but it does not enhance the intrinsic generation capability of autoregressive models. The performance gain is model-dependent. If the base model is trained with techniques that promote diverse token distributions, such as noise injection during training, or is well-trained on large-scale datasets, the improvement tends to be limited. Moreover, for weak base models, such as LlamaGen Stage 2, the method may offer little or no performance gain.
\section{Additional visual comparison}
\label{sec_add_visual_comparison}

\begin{figure*}[h]
\setlength{\abovecaptionskip}{0.1cm}
\setlength{\belowcaptionskip}{0.1cm}
\begin{center}
\includegraphics[width=0.9\textwidth]{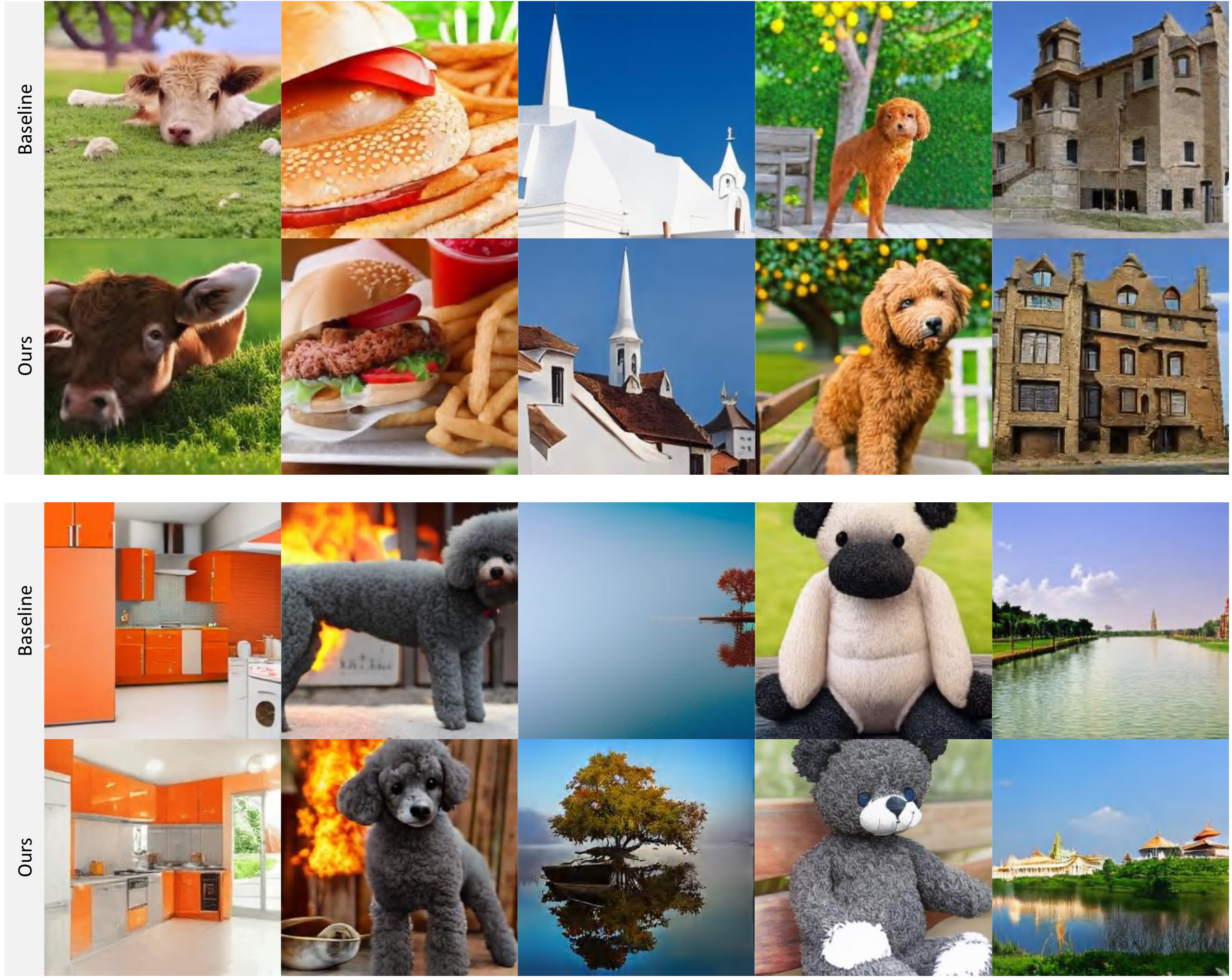}
\end{center}
\caption{
Visualization of LlamaGen.
}
\vspace{-0.3cm}
\label{fig_supp_add_visual_llamagen}
\end{figure*}

\begin{figure*}[h]
\setlength{\abovecaptionskip}{0.1cm}
\setlength{\belowcaptionskip}{0.1cm}
\begin{center}
\includegraphics[width=1.0\textwidth]{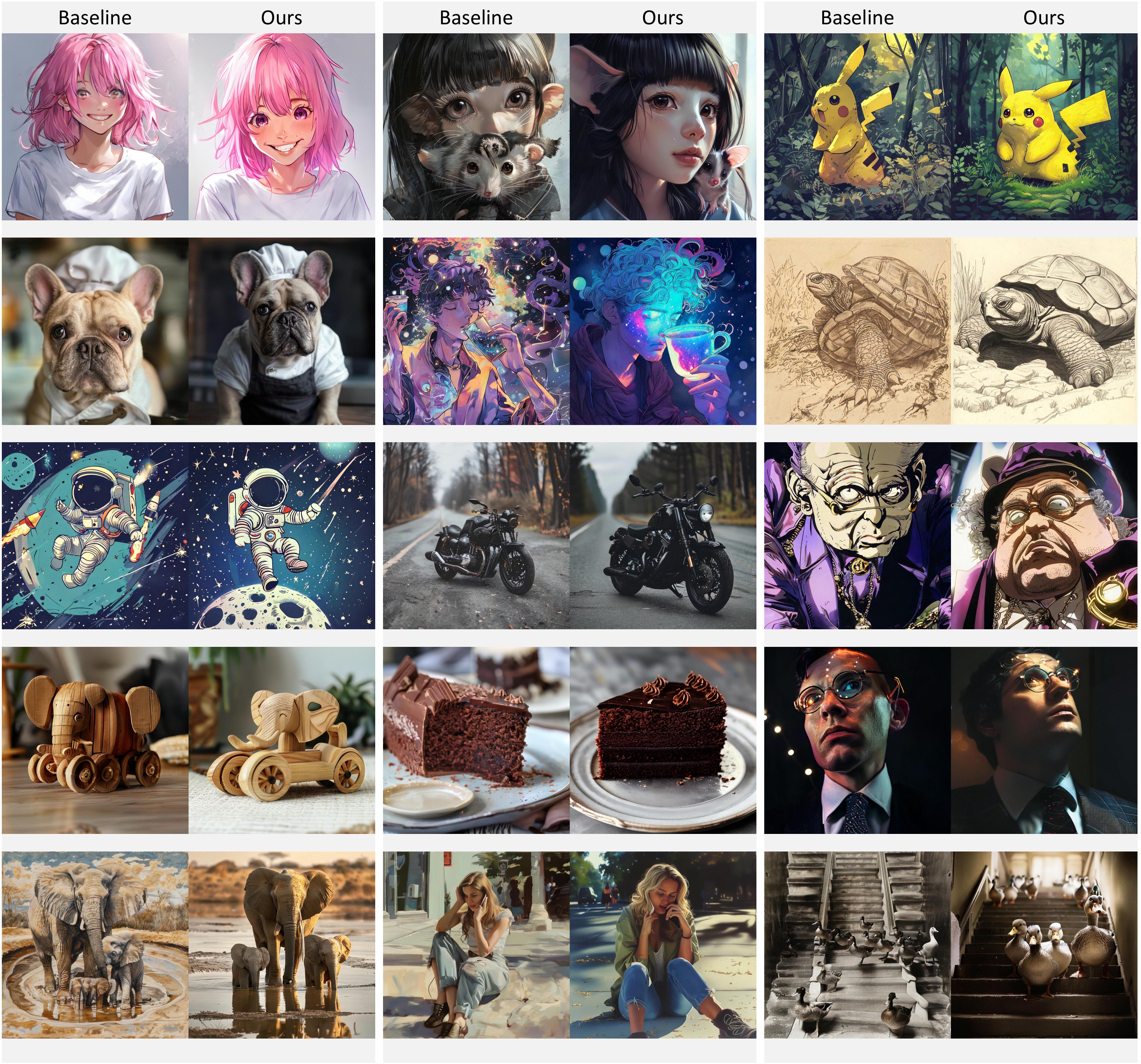}
\end{center}
\caption{
Visualization of Lumina-mGPT.
}
\vspace{-0.3cm}
\label{fig_supp_add_visual_mgpt}
\end{figure*}

\begin{figure*}[h]
\setlength{\abovecaptionskip}{0.1cm}
\setlength{\belowcaptionskip}{0.1cm}
\begin{center}
\includegraphics[width=1.0\textwidth]{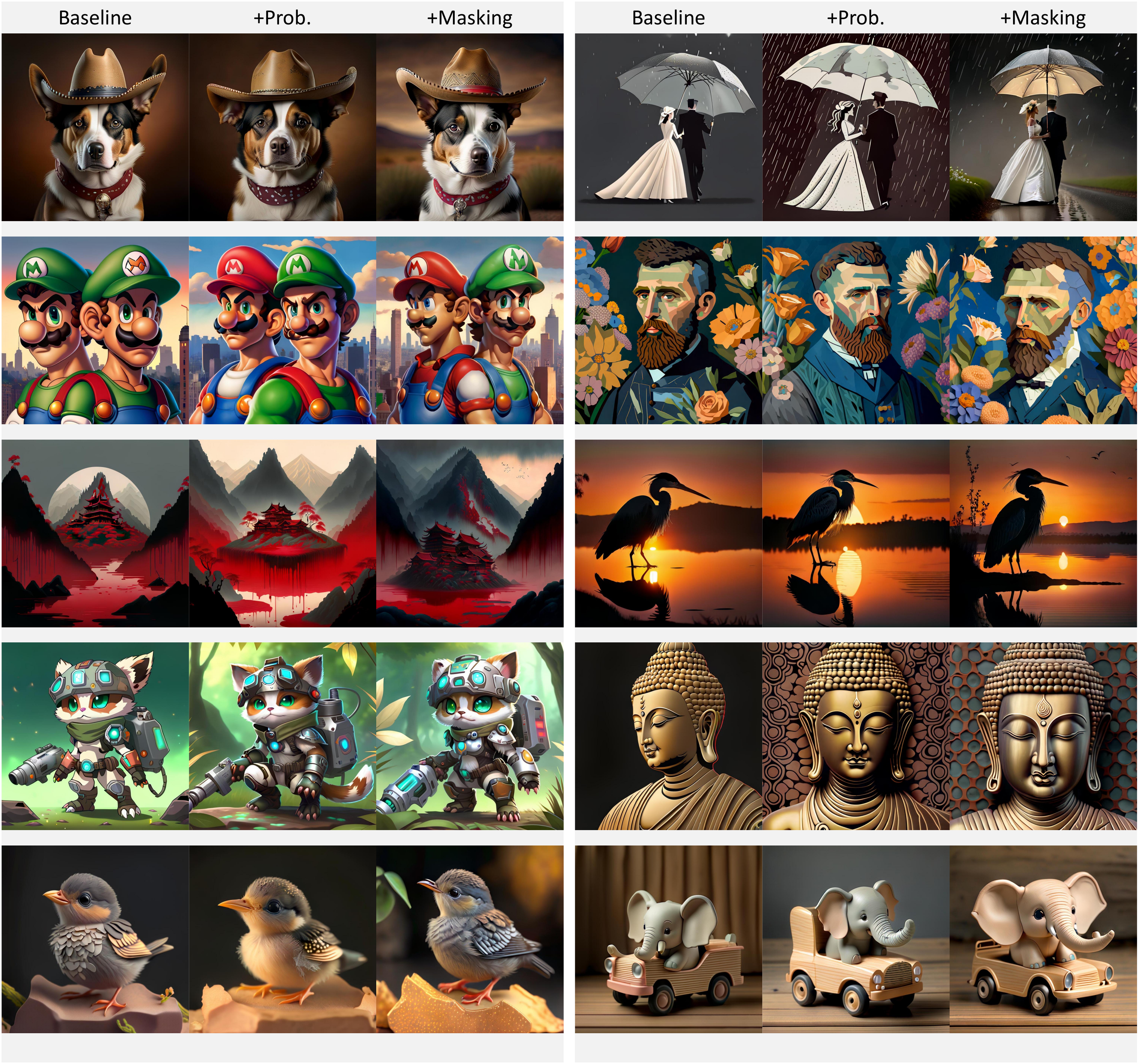}
\end{center}
\caption{
Visualization of Meissonic.
}
\vspace{-0.3cm}
\label{fig_supp_add_visual_meissonic}
\end{figure*}

\begin{figure*}[h]
\setlength{\abovecaptionskip}{0.1cm}
\setlength{\belowcaptionskip}{0.1cm}
\begin{center}
\includegraphics[width=1.0\textwidth]{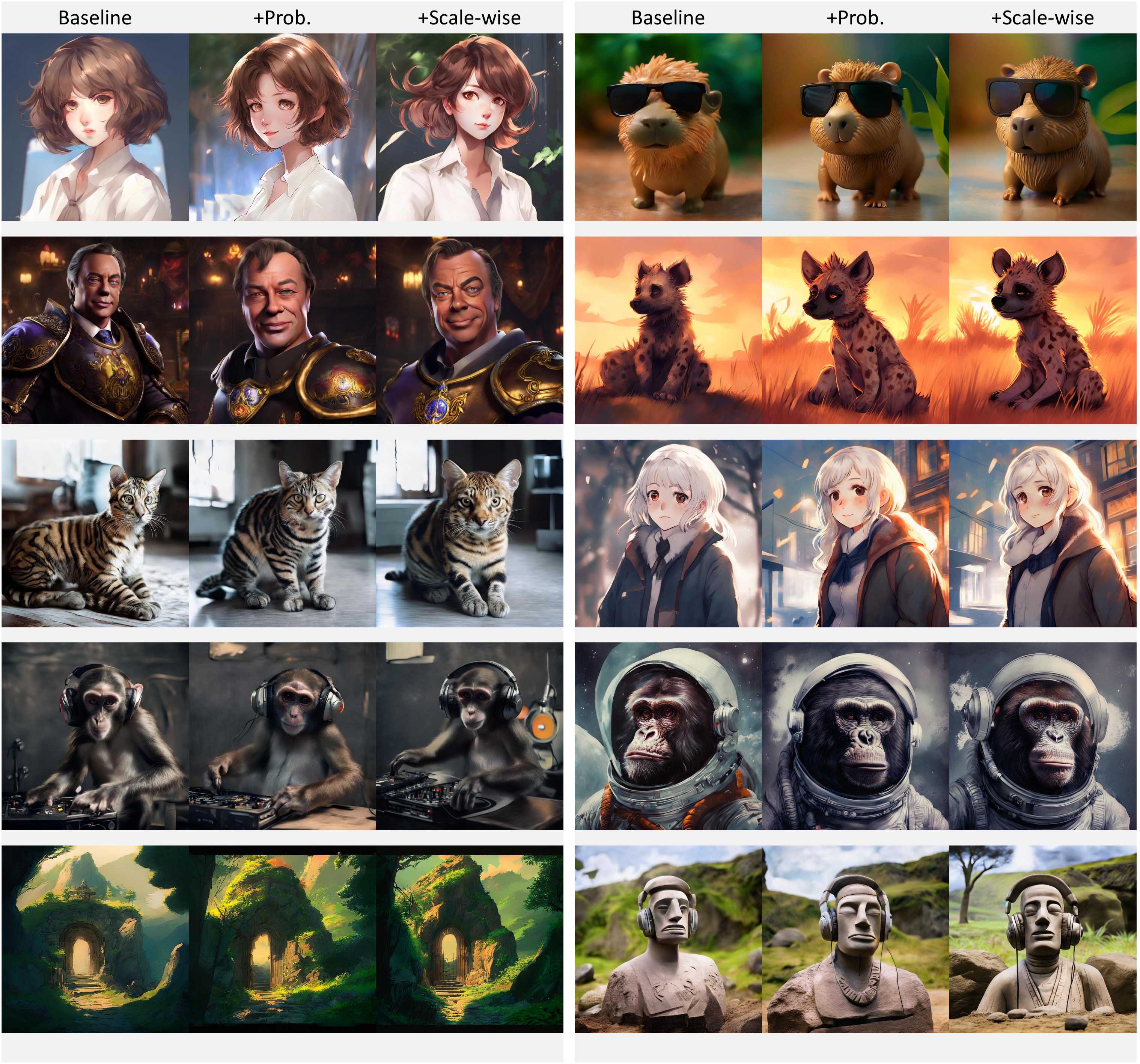}
\end{center}
\caption{
Visualization of STAR.
}
\vspace{-0.3cm}
\label{fig_supp_add_visual_star}
\end{figure*}

\begin{figure*}[h]
\setlength{\abovecaptionskip}{0.1cm}
\setlength{\belowcaptionskip}{0.1cm}
\begin{center}
\includegraphics[width=1.0\textwidth]{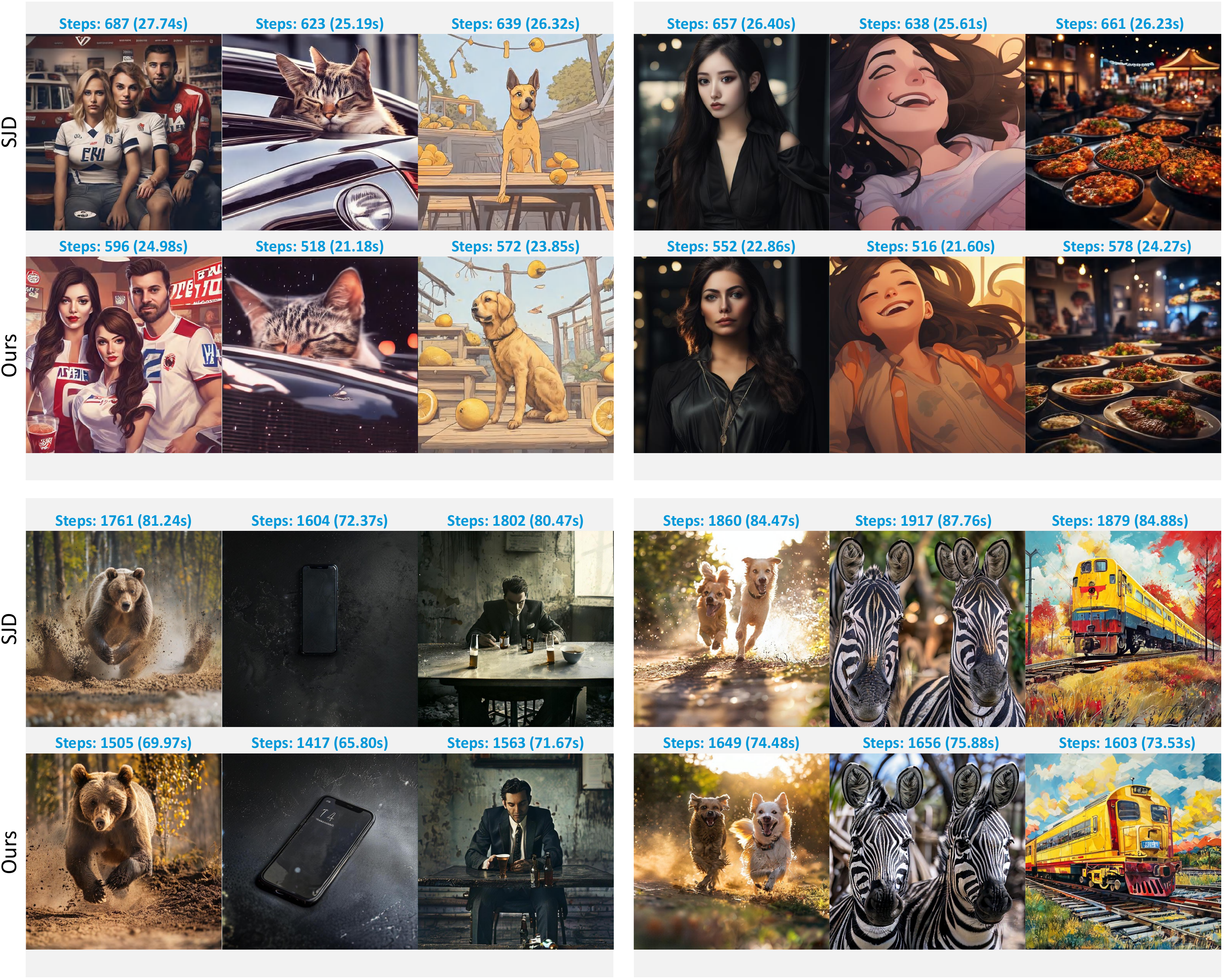}
\end{center}
\caption{
Visualization of AR acceleration.
}
\vspace{-0.3cm}
\label{fig_supp_add_visual_acc}
\end{figure*}

Due to space constraints in the main paper, we did not provide additional visualizations. Here, we include further results illustrating the entropy-aware sampling behavior for LlamaGen, Lumina-mGPT, Meissonic, and STAR, as well as acceleration visualizations for LlamaGen and Lumina-mGPT (see Fig.~\ref{fig_supp_add_visual_llamagen}–Fig.~\ref{fig_supp_add_visual_acc}).




\end{document}